\pdfoutput=1

\documentclass[11pt]{article}

\usepackage{acl}

\usepackage{times}
\usepackage{latexsym}

\usepackage[T1]{fontenc}

\usepackage[utf8]{inputenc}

\usepackage{microtype}

\usepackage{inconsolata}

\usepackage{graphicx}

\usepackage{color, colortbl} %
\usepackage{fancyvrb}
\usepackage{tcolorbox}
\usepackage{booktabs}
\usepackage{multirow}
\usepackage{amssymb}
\usepackage{soul}
\usepackage[normalem]{ulem} %
\definecolor{sand}{HTML}{E1DAAE}
\definecolor{blue}{HTML}{058ED9}
\definecolor{orange}{HTML}{FF934F}
\definecolor{red}{HTML}{CC2D35}
\colorlet{red10}{red!10}
\colorlet{red20}{red!20}
\colorlet{red30}{red!30}
\colorlet{red40}{red!40}
\colorlet{red50}{red!50}
\colorlet{red60}{red!60}
\colorlet{red70}{red!70}
\colorlet{red80}{red!80}
\colorlet{red90}{red!90}
\colorlet{red100}{red!100}
\colorlet{orange10}{orange!10}
\colorlet{orange20}{orange!20}
\colorlet{orange30}{orange!30}
\colorlet{orange33}{orange!33}
\colorlet{orange40}{orange!40}
\colorlet{orange50}{orange!50}
\colorlet{orange60}{orange!60}
\colorlet{orange70}{orange!70}
\colorlet{orange80}{orange!80}
\colorlet{orange90}{orange!90}
\colorlet{orange100}{orange!100}
\newcommand{\hluline}[2]{%
    \setlength{\fboxsep}{1pt}%
    \colorbox{#1}{\uline{\textcolor{black}{#2}}}%
}
\usepackage{pifont}
\newcommand{\cmark}{\ding{51}}%
\newcommand{\xmark}{\ding{55}}%
\usepackage{caption}
\usepackage{subcaption}
\usepackage{footmisc} %
\usepackage{CJK}
\usepackage{mdframed}
\usepackage{listings}
\usepackage{hyperref}
\usepackage{tablefootnote}

\title{What's Mine becomes Yours: 
Defining, Annotating and Detecting 
Context-Dependent Paraphrases in News Interview Dialogs}

\author{
    Anna Wegmann\textsuperscript{1}, Tijs van den Broek\textsuperscript{2} and   Dong Nguyen\textsuperscript{1} \\
    \textsuperscript{1}Utrecht University, Utrecht, The Netherlands \\
    \textsuperscript{2}Vrije Universiteit Amsterdam, Amsterdam, The Netherlands \\
    \texttt{\{a.m.wegmann, d.p.nguyen\}@uu.nl}, \texttt{t.a.vanden.broek@vu.nl}
  }

\begin{document}
\maketitle
\begin{abstract}
    Best practices for high conflict conversations like counseling or customer support almost always include recommendations to paraphrase the previous speaker. Although paraphrase classification has received widespread attention in NLP, paraphrases are usually considered independent from context, and common models and datasets are not applicable to dialog settings. In this work, we investigate paraphrases across turns in dialog (e.g., Speaker 1: ``That book is mine.'' becomes Speaker 2: ``That book is yours.''). We provide an operationalization of context-dependent paraphrases, and develop a  training for crowd-workers to classify paraphrases in dialog. We introduce \texttt{ContextDeP}, a dataset with utterance pairs from NPR and CNN news interviews annotated for context-dependent paraphrases. To enable analysis on label variation, the dataset contains 5,581 annotations on 600 utterance pairs. We present promising results with in-context learning and with token classification models for automatic paraphrase detection in dialog. %
\end{abstract}

\section{Introduction} \label{sec:introduction}

    Repeating or paraphrasing what the previous speaker said has time and time again been found to be important in human-to-human or human-to-computer dialogs: It encourages elaboration and introspection in counseling \citep{Rogers-Listening, miller2012motivational, hill1992overview, shah2022modeling}, 
    can help deescalate conflicts in crisis negotiations \citep{vecchi_hostage-negotation,voss2016negotiating,vecchi2019negotiating}, 
    can have a positive impact on  relationships \citep{weger2010active, RoosFeelingHeard},
    can increase the perceived response quality of dialog systems \citep{weizenbaum1966eliza, dieter-etal-2019-mimic}
    and generally provides tangible understanding-checks to {ground} what both speakers agree on \citep{clark1996using, jurafsky2019speech}.

    \begin{figure}
        \small
        \textbf{Guest: }
        \colorlet{orange0}{orange!0}\sethlcolor{orange0}\hl{
        And people always prefer, of course, to see the pope as the principal celebrant of the mass. So that's good. That'll be tonight. And it will be his 26th mass and it will be the 40th or, rather, the 30th time that this is offered in round the world transmission. And } \sethlcolor{orange100}\hl{it will be }\hluline{orange100}{my 20th time in doing it as a} \hluline{orange100}{television commentator}\colorlet{orange33}{orange!33.33333333333333}\sethlcolor{orange33}\hl{ from Rome} \colorlet{orange0}{orange!0}\sethlcolor{orange0}\hl{so. } 
         \\ 
        \textbf{Host: }
        \colorlet{orange0}{orange!0}\sethlcolor{orange0}\hl{Yes, } \colorlet{orange100}{orange!100.0}\sethlcolor{orange100}\hluline{orange100}{you've been doing this for a while now.} 
        \caption{\textbf{Context-Dependent Paraphrase in a News Interview.} The interview host %
        {paraphrases} part of %
        {the guest's utterance}.  It is only a paraphrase in the current context (e.g., \textit{doing something 20 times} and \textit{doing something for a while} are not generally synonymous). Our annotators provide word-level \hl{highlighting}. The color's intensity shows the share of annotators that selected the word. Here, most annotators selected the same text spans, some included ``from Rome'' as part of what is paraphrased by the host. We \uline{underline} the {paraphrase} identified by our fine-tuned \texttt{DeBERTa} token classifier.
        }
        \label{fig:ex-understanding}
        \label{fig:figure-1}
    \end{figure}

        \begin{table*}[t]
            \centering \footnotesize 
             \begin{tabular}{ l   c c  p{105mm} p{5mm}} 
             \toprule
                 & \multicolumn{2}{c}{\textbf{Agreement}} & \multicolumn{2}{c}{\textbf{Single Example with High Variation}} \\
                \textbf{Dataset} & \textbf{Acc.} & \textbf{$\alpha$} &  \textbf{Shortened  Example} & \textbf{Vote} \\ 
            \midrule
                \vspace*{-1.2\baselineskip}BAL\-ANCED & 0.71 & 0.32 & 
                    \parbox{105mm}{
                            \textbf{Guest: } [...]
                            \colorlet{orange44}{orange!44.44444444444444}\sethlcolor{orange44}\hl{Maybe } \colorlet{orange66}{orange!66.66666666666666}\sethlcolor{orange66}\hl{the } \colorlet{orange77}{orange!77.77777777777779}\sethlcolor{orange77}\hl{money will help. } 
                             \\ 
                            \textbf{Host: }
                            \colorlet{orange88}{orange!88.88888888888889}\sethlcolor{orange88}\hl{It } \colorlet{orange100}{orange!100.0}\sethlcolor{orange100}\hl{can't hurt, } \colorlet{orange22}{orange!22.22222222222222}\sethlcolor{orange22}\hl{let's put it that way. } 
                    }  & 9/20 \\ \midrule
            RAN\-DOM & 0.72 & 0.23 & \parbox{105mm}{
                    \textbf{G:}
                    So both parties agree that \colorlet{orange14}{orange!14.285714285714285}\sethlcolor{orange14}\hl{we need to } \colorlet{orange28}{orange!28.57142857142857}\sethlcolor{orange28}\hl{stop horrific acts of } \colorlet{orange57}{orange!57.14285714285714}\sethlcolor{orange57}\hl{violence against animals. } %
                    \colorlet{orange14}{orange!14.285714285714285}\sethlcolor{orange14}\hl{But } \colorlet{orange28}{orange!28.57142857142857}\sethlcolor{orange28}\hl{everyone is standing behind this. It is time to } \colorlet{orange71}{orange!71.42857142857143}\sethlcolor{orange71}\hl{stop horrific acts of } \colorlet{orange85}{orange!85.71428571428571}\sethlcolor{orange85}\hl{brutality on animals. } 
                     \\ 
                    \textbf{H: }
                    Britain's Queen Elizabeth's senior dresser writes "If her majesty is due to attend an engagement in particularly cold weather from 2019 onwards fake fur will be used to make sure she stays warm." 
                    it's a very stark example of  \colorlet{orange14}{orange!14.285714285714285}\sethlcolor{orange14}\hl{a monarch following public } \colorlet{orange28}{orange!28.57142857142857}\sethlcolor{orange28}\hl{opinion in the U.K. which is moving away from fur and it } \colorlet{orange42}{orange!42.857142857142854}\sethlcolor{orange42}\hl{very much } \colorlet{orange57}{orange!57.14285714285714}\sethlcolor{orange57}\hl{embraces } \colorlet{orange85}{orange!85.71428571428571}\sethlcolor{orange85}\hl{prevention of } \colorlet{orange100}{orange!100.0}\sethlcolor{orange100}\hl{cruelty to the animals. } 
                    }
                     & 7/15 \\ \midrule
            PARA & 0.65 & 0.19 & \parbox{105mm}{
                    \textbf{G: }
                    [...]
                    \colorlet{orange12}{orange!12.5}\sethlcolor{orange12}\hl{it could be programmed in. } {But again, } \colorlet{orange75}{orange!75.0}\sethlcolor{orange75}\hl{you'd have to set that up } \colorlet{orange87}{orange!87.5}\sethlcolor{orange87}\hl{as part of your } \colorlet{orange100}{orange!100.0}\sethlcolor{orange100}\hl{flight plan. } 
                     \\ 
                    \textbf{H: }
                    \colorlet{orange0}{orange!0}\sethlcolor{orange0}\hl{So } \colorlet{orange37}{orange!37.5}\sethlcolor{orange37}\hl{you'd have to say } \colorlet{orange50}{orange!50.0}\sethlcolor{orange50}\hl{I'm going to } \colorlet{orange62}{orange!62.5}\sethlcolor{orange62}\hl{drop } \colorlet{orange50}{orange!50.0}\sethlcolor{orange50}\hl{to 5,000 feet, } \colorlet{orange62}{orange!62.5}\sethlcolor{orange62}\hl{then go back up } \colorlet{orange50}{orange!50.0}\sethlcolor{orange50}\hl{to 35,000 feet, } \colorlet{orange12}{orange!12.5}\sethlcolor{orange12}\hl{and } \colorlet{orange62}{orange!62.5}\sethlcolor{orange62}\hl{you would have had to have done that at the beginning. } 
                } & 8/15 \\ %

             \bottomrule
        \end{tabular}
        \caption{\textbf{Agreement Scores as an Indicator of Plausible Variation.} \label{tab:annotation-difficulties-examples} \label{tab:agreement}
            For each dataset, we display 
             the ``accuracy'' with the majority vote (Acc.) which is the mean overlap of a rater's classification with the majority vote classification excluding the current rater and 
              \citet{krippendorff2018content}'s alpha ($\alpha$) for the binary classifications by all raters over all pairs.    
            The relatively low K's $\alpha$ scores can be explained by pairs where either label is plausible. We display such an example for each dataset with the share of annotators classifying it it as a paraphrase  (Vote). %
        }
        \end{table*}

    Fortunately, in NLP, paraphrases have received wide-spread attention: Researchers have created numerous paraphrase datasets \citep{dolan-brockett-2005-automatically,zhang-etal-2019-paws,dong-etal-2021-parasci,kanerva_finish-paraphrase}, 
    developed methods to automatically identify paraphrases \citep{zhang-etal-2019-paws, wei2021finetuned, zhou2022paraphrase},
    and used paraphrase datasets to train semantic sentence representations \citep{reimers-gurevych-2019-sentence,gao-etal-2021-simcse} 
    and benchmark LLMs \citep{wang-etal-2018-glue, bigbench2022}. 
    However, most previous work (1) has focused on context-independent paraphrases, i.e., texts that are semantically equivalent independent from the given context, and has not investigated the automatic detection of paraphrases across turns in dialog, (2) has classified paraphrases at the level of full texts even though paraphrases often only occur in portions of larger texts (see also Figure \ref{fig:figure-1}), (3) uses a small number of 1--3 annotations per paraphrase pair \citep{dolan-brockett-2005-automatically, kanerva_finish-paraphrase}, (4) only annotate text pairs that are ``likely'' to include paraphrases using heuristics such as lexical similarity \citep{dolan-brockett-2005-automatically}, although, especially for the dialog setting, we can not expect lexical similarity to be high for all or even most paraphrase pairs (e.g., the pair in Figure~\ref{fig:figure-1} only overlaps in two words) and (5) either use short annotation instructions \cite{dolan-brockett-2005-automatically} that rely on annotator intuitions or long and complex instructions \cite{kanerva_finish-paraphrase} that limit the total number of annotators.

    We address all five limitations with this work.
    First, we are, to the best of our knowledge, the first to focus on operationalizing, annotating and automatically detecting \textbf{context-dependent paraphrases across turns in dialog}. Dialog is a setting that is uniquely sensitive to context \citep{grice1957meaning, grice1975logic,davis2002meaning}, e.g., ``doing this for a while now'' and ``20th time [...] as a television commentator'' in Figure \ref{fig:figure-1} are not generally semantically equivalent. 
    Second, instead of classifying whether two complete texts A and B are paraphrases of each other, %
    we focus on {classifying} whether there exists a selection of a text B that {paraphrases} a selection of a text A, and \textbf{identifying the text spans that constitute the paraphrase pair} (e.g., Figure~\ref{fig:figure-1}). 
    Third, we collect a \textbf{larger number of annotations} of up to 21 per item in line with typical efforts to address plausible human label variation \citep{nie-etal-2020-learn, sap-etal-2022-annotators}. Even though context-dependent paraphrase identification in dialog might at first seem straight forward with a clear ground truth, similar to other ``objective'' tasks in NLP \citep{uma2021learning}, human annotators (plausibly) disagree on labels \citep{dolan-brockett-2005-automatically, kanerva_finish-paraphrase}. For example, consider the first text pair in Table \ref{tab:annotation-difficulties-examples}. ``[The money] can't hurt'' can be interpreted in at least two different ways: as a statement with approximately the same meaning as ``the money will help'' or as an opposing statement meaning the money actually won't help but at least ``It can't hurt'' either.
    Fourth, instead of using heuristics to select text pairs for annotations, we choose a dialog setting where paraphrases are relatively likely to occur: transcripts of \textbf{NPR and CNN news interviews} \citep{zhu-etal-2021-mediasum} since in {(news) interviews} paraphrasing or more generally active listening is encouraged \citep{clayman2002news, hight2002tragedies, sedorkin2023interviewing}. While the interview domain shows some unique characteristics limiting generalizability (e.g., hosts using paraphrases to simplify the guest's statements for the audience), the interview domain is is suitable to demonstrate our new task and includes a diverse set of topics and guests.
    Fifth, we develop an annotation procedure that goes beyond relying on intuitions and is scalable to a large number of annotators: an accessible \textbf{example-centric, hands-on, 15-minute training} before annotation.

    In short, we operationalize {context-dependent paraphrases} in dialog with a definition and 
    an iteratively developed hands-on training for annotators. Then, annotators classify paraphrases and identify the spans of text that constitute the paraphrase. %
    We release \texttt{ContextDeP} (\uline{Context}-\uline{De}pendent \uline{P}araphrases in news interviews),
    a dataset with  5,581 annotations on 600 utterance pairs from NPR and CNN news interviews. %
    We use in-context learning (ICL) with  generative models like \texttt{Llama 2} or \texttt{GPT-4} and fine-tune a \texttt{DeBERTa} token classifier to detect paraphrases in dialog. We reach promising results of F1 scores from $0.73$ to $0.81$.
    Generative models perform better at classification, while the token classifier provides text spans without parsing errors.
    We hope to advance dialog based evaluations of LLMs and the reliable detection of paraphrases in dialog. 
    Code\footnote{\url{https://github.com/nlpsoc/Paraphrases-in-News-Interviews}}, annotated data\footnote{\url{https://huggingface.co/datasets/AnnaWegmann/Paraphrases-in-Interviews}}$^{,}$\footnote{This is in line with the license from the original data publication \cite{zhu-etal-2021-mediasum}.} and the trained model\footnote{\url{https://huggingface.co/AnnaWegmann/Highlight-Paraphrases-in-Dialog}} are publicly available for research purposes.

    \begin{table}[t]
            \centering \footnotesize 
             \begin{tabular}{%
                p{15mm}  p{50mm}} 
             \toprule
              \textbf{What?} &  \textbf{Shortened  Examples} \\ 
            \midrule
             \multirow{5}{15mm}{{\textbf{Clear Con\-tex\-tu\-al Equivalence $\subseteq$ CP}
                    }
                }
                & 
                    \parbox{50mm}{
                            Guest: {I} {know they are} {cruel}. \\ 
                            Host: {You} {know they are} cruel.} \\\cmidrule{2-2}
                    & \parbox{50mm}{
                            G: {We} have been the punching bag of the president. \\ 
                            H: The president has been using {Chicago} as a punching bag.
                            } \\
                \cmidrule{2-2}   
                \vspace{-\baselineskip} \multirow{1}{15mm}{ \textbf{App\-rox\-i\-mate Con\-tex\-tual Equivalence $\subseteq$ CP}} 
                    & \parbox{50mm}{
                            G: I'm like, "Fortnite", what is that? I don't even know what it is --\\
                            {H:} 
                            So, you weren't even familiar?} \\\cmidrule{2-2}
                    & \parbox{50mm}{
                            G: My wife is {going through the same thing} herself. \\
                            H: She's {also looking for work}.} \\
             \bottomrule
            \end{tabular}
            \caption{\textbf{Contextual Paraphrases (CP).}      \label{tab:CP-examples} 
                We include text spans ($\subseteq$ CP) that range from clear to approximate equivalence for the given context. Few examples are very clear. Deciding between approximate equivalence and non-equivalence turns out to be a difficult task. In our dataset, annotator agreement scores can be used as a proxy for the ambiguity of an item.
            }
        \end{table}
        
    \begin{table}[t]
            \centering \footnotesize 
             \begin{tabular}{%
                p{13mm} p{53mm}} 
             \toprule
              \textbf{What?} &  \textbf{Shortened  Example} \\
                \midrule
                    \vspace{-3.15\baselineskip}\multirow{1}{13mm}{\textbf{Ad\-di\-tion\-al Con\-clu\-sions or Facts $\nsubseteq$ CP}}
                    &
                         \parbox{53mm}{
                            Guest: If you're not in our country, there are no constitutional protections for you.\\
                            Host: So, {you don't have a problem with} Facebook giving the government access to 
                            the private accounts of people applying to enter the U.S.?
                        }  \\ %
                \midrule
                \vspace*{-2.2\baselineskip} \textbf{Isolated Equivalence $\nsubseteq$ CP} \vspace*{-1\baselineskip} & 
                         \parbox{53mm}{ 
                                G: There are militant groups out there firing against \uline{the military}. \\
                                H: Why did \uline{the army} decide today to move in and clear out the camp? } \vspace*{-0.5\baselineskip} \\ %
             \bottomrule
            \end{tabular}
            \caption{\textbf{Non-Paraphrases in Dialog.} \label{tab:NCP-examples} We do not include text pairs ($\nsubseteq$ CP) that are  semantically related but where the second speaker does not actually rephrase a point the first speaker makes. Frequent cases are text spans that might  only be considered approximately equivalent when taken out of context (\uline{underlined}) %
            and pairs that have too distant meanings, for example, when the interviewer continues with the same or a related topic but adds further-reaching conclusions or new facts.
            }
        \end{table}

\section{Related Work}

    Paraphrases have most successfully been classified by 
    encoder architectures with fine-tuned classification heads \citep{zhang-etal-2019-paws,wahle-etal-2023-paraphrase} and more recently using in-context learning with generative models like \texttt{GPT-3.5} and \texttt{Llama~2} \citep{wei2021finetuned,wang-etal-2022-super,wahle-etal-2023-paraphrase}. 
    To the best of our knowledge, only \citet{wang-etal-2022-paratag} go beyond classifying paraphrases at the complete sentence level. They use a \texttt{DeBERTa} token classifier to highlight text spans that are not part of a paraphrase, i.e., the reverse of our task.

     Paraphrase taxonomies commonly go beyond binary classifications to make more fine-grained distinctions between paraphrase types, often including considerations w.r.t. the context of the text pairs. 
     \citet{bhagat-hovy-2013-squibs} and \citet{kovatchev-etal-2018-etpc} describe substitutions and other lexical operations that result in paraphrases in a given sentential context. 
     \citet{shwartz-dagan-2016-adding} show that context information can reverse semantic relations between phrases. 
     \citet{vila2014paraphrase} discuss text pairs that are equivalent when one presupposes encyclopedic or situational knowledge (e.g., referents or intentions\footnote{cases like `Close the door please'' and ``There is air flow''}), but exclude them as non-paraphrases. 
    Further, to the best of our knowledge, most previous work annotate sentence pairs without considering the  document context, with \citet{kanerva_finish-paraphrase} being the only exception, and no previous work looking at detecting paraphrases in dialog.

    Dialog act taxonomies aim to classify the communicative function of an utterance in dialog and commonly include acts such as \texttt{Summarize/Reformulate}  \citep{stolcke-etal-2000-dialogue,DAMSL-core1997}.
        However, generally, communicative function can be orthogonal to meaning equivalence. For example, the paraphrase from Table \ref{tab:CP-examples} ``So you weren't even familiar?'' would probably be a \texttt{Declarative Yes-No-Question} dialog act \citep{stolcke-etal-2000-dialogue},
        while the non-paraphrase ``So you don't have a problem with ... ?'' in Table \ref{tab:NCP-examples} would also be a \texttt{Declarative Yes-No-Question}.
    We see paraphrase detection in dialog as more elementary and complementary to investigating communicative function of utterances.

    \section{Context-Dependent Paraphrases in Dialog}
    \label{sec:paraphrase-definition}

    In NLP, paraphrases typically are pairs of text that are {approximately equivalent} in meaning \citep{bhagat-hovy-2013-squibs}, since full equivalence usually only applies for practically identical strings \citep{bhagat-hovy-2013-squibs, dolan-brockett-2005-automatically}
    -- with
    some scholars even claiming that different sentences can never be fully equivalent in meaning \citep{hirst2003paraphrasing, clark1992conventionality, bolinger1974meaning}. 
    The field of NLP has mostly focused on paraphrases that are \textit{context-independent}, i.e., approximately equivalent without considering a given context \citep{dolan-brockett-2005-automatically,wang-etal-2018-glue,zhang-etal-2019-paws}. 
    Some studies have operationalized paraphrases using more fine-grained taxonomies, where context is sometimes considered \citep{bhagat-hovy-2013-squibs, vila2014paraphrase, kovatchev-etal-2018-etpc}. However, only a few datasets include such paraphrases \citep{kovatchev-etal-2018-etpc, kanerva_finish-paraphrase} and to the best of our knowledge none that focus on context-dependent paraphrases or dialog data.

    We define a \textbf{context-dependent paraphrase} as two text excerpts that are at least approximately equivalent in meaning {in a given situation} but not necessarily in all non-absurd situations.\footnote{definition combines elements from \citet{kanerva2021annotation} and \citet{bhagat-hovy-2013-squibs}}
    For example, consider the first exchange in Table~\ref{tab:CP-examples}. In this situation, ``I'' uttered by the first speaker and ``You'' uttered by the second speaker are clearly signifying the same person. However, if uttered by the same speaker ``I'' and ``you'' probably do not signify the same person. The text pair in Table~\ref{tab:CP-examples} is thus equivalent in at least one but not in all non-absurd situations.
        The text excerpts forming context-dependent paraphrases do not have to be complete utterances. In many cases they are portions of utterances, see  highlights in Figure~\ref{fig:ex-understanding}. 
    Note that in dialog, the second speaker should rephrase part of the first speaker's point in the given situation (\textit{context} condition) and not just talk about something semantically related (\textit{equivalence} condition).  
    
    Context-dependent paraphrases range from clear (first example in Table~\ref{tab:CP-examples}) to approximate contextual equivalence (last example in Table~\ref{tab:CP-examples}). When the guest says ``My wife is going through the same thing'', it seems reasonable to assume that the host is using contextual knowledge %
    to infer that ``the same thing'' and ``looking for a job'' are equivalent for the given exchange. 
    Even though in this last example the meaning of the two utterances could also be subject to different interpretations,  %
    we still consider such cases to be context-dependent paraphrases for two reasons: (1) similar to findings in context-independent paraphrase detection, limiting ourselves to very clear cases would mostly result in uninteresting, practically identical strings and (2) we ultimately want to identify paraphrases in human dialog, which is full of implicit contextual meaning \citep{grice1957meaning, grice1975logic,davis2002meaning}.

    We specifically exclude common cases of disagreements between annotators\footnote{
        derived from pilot studies, see also App. \ref{app:annotator-training-development} and specifically App. Table \ref{ex:disagreements}}
    that we consider not to be context-dependent paraphrases in dialog, see Table \ref{tab:NCP-examples}. 
    First, we exclude text spans that might be considered approximately equivalent when they are looked at in isolation but do not represent a paraphrase of the guest's point in the given situation (e.g., ``the military'' and ``the army'' in Table~\ref{tab:NCP-examples}).
    Second, we exclude text pairs that diverge too much from the original meaning when the second speaker adds conclusions, inferences or new facts. In an interview setting, journalists make use of different question types and communication strategies relating to their agenda \citep{clayman2002news} that can sometimes seem like paraphrases. For example in Table \ref{tab:NCP-examples}, the host's question ``So, you ...?'' could be read as a paraphrase with the goal of checking understanding with the guest. %
    However, it is more likely to be a declarative conclusion that goes beyond what the guest said.

    \label{sec:paraphrase-operationalization}

\section{Dataset} \label{sec:mediasum}

    Generally, people do not paraphrase each other in every conversation.
    We focus on the news interview setting, because paraphrasing, or more generally active listening, is a common practice for journalists \citep{clayman2002news, hight2002tragedies, sedorkin2023interviewing}. We therefore also only consider whether the journalist (the interview host) paraphrases the interview guest and not the other way around. We use \citet{zhu-etal-2021-mediasum}'s \textit{MediaSum} corpus which consists of over 450K news interview transcripts and their summaries from 1999--2019 NPR %
    and 2000--2020 CNN interviews.\footnote{Released for research purpose, see \url{https://github.com/zcgzcgzcg1/MediaSum?tab=readme-ov-file}.} %

    \subsection{Preprocessing}\label{sec:preprocessing}

    We only include two-person interviews, i.e., a conversation between an interview host and a guest. %
    We remove interviews with fewer than four turns, utterances that only consist of two words or of more than 200 words, and the first and last turns of interviews (often welcoming addresses and goodbyes). 
    Overall, this leaves 34,419 interviews with 148,522 (guest, host)-pairs. See App.~\ref{app:data-2person-filter} for details.

    \subsection{Data Samples for Annotation} \label{sec:candidate-selection}
    \label{sec:annotation-sets}

        Even though paraphrases are relatively likely in the news interview setting, most randomly sampled text pairs still do not include paraphrases. %
        To distribute annotation resources to text pairs that are likely to be paraphrase, previous work usually selects pairs based on heuristics like textual similarity features, e.g., word overlap, edit distance, or semantic similarity \citep{dolan-brockett-2005-automatically, su-yan-2017-cross, dong-etal-2021-parasci}. 
        However, these approaches are systematically biased towards selecting more obvious, often lexically similar text pairs, possibly excluding many context-dependent paraphrases. For example, the guest and host utterance in Figure~\ref{fig:ex-understanding} have varying lengths, only overlap in three words and have a semantic similarity score of only 0.13\footnote{\label{sbert} using cosine-similarity and encodings from \url{https://huggingface.co/sentence-transformers/all-mpnet-base-v2}}. Similar to \citet{kanerva_finish-paraphrase}, we instead use a manual selection of promising text pairs for annotation: We (1) randomly sample a  set of text pairs and (2) manually classify at each of them to (3) select three sets of text pairs that vary in their paraphrase distribution for the more resource-intensive crowd-sourced annotations: the RANDOM, BALANCED and PARA set.

    \begin{table}[t]
        \centering \small
         \begin{tabular}{l c r r } 
         \toprule
          \textbf{Dataset} & \textbf{size} & \textbf{\# paraphrases} & \textbf{\# anns/item} \\ 
        \midrule             
            BALANCED & 100 & 54 & 20.1 \\ %
            RANDOM & 100 & 13 & 5.7 \\ %
            PARA & 400 & 254 & 7.5 \\ %
        \midrule
            Total & 600 & 321 & 9.3 \\ %
        \bottomrule
        \end{tabular}
        \caption{\textbf{Dataset Statistics.} \label{tab:label-dist}
            For each dataset, we display the size, the number of paraphrases according to the majority vote and the average annotations per text pair. 
        }
    \end{table}

        \textbf{Lead Author Annotation.} We shuffle and uniformly sample 1,304 interviews. For each interview, we sample a maximum of 5 consecutive (guest, host)-pairs.
            To select promising paraphrase candidates, the lead author then manually classifies all 4,450 text pairs %
            as paraphrases vs. non-paraphrases (see App. \ref{app:paraphrase-candidates} for details).\footnote{
                After experimenting with crowd-workers, having a first pass for selection done by one of our team seemed the best considering cost-performance trade-offs.
            } 
            In total, about 14.9\% of the sampled text pairs are classified as %
            paraphrases by the lead author. %
            On a random set of 100 (guest, host)-pairs (RANDOM), we later compare the lead author's classifications with the crowd-sourced  paraphrase classifications (see App. \ref{app:lead-crowd}). 89\% of the lead author's classifications are the same as the crowd majority. 
            Note that the lead author's classifications do not affect the quality of the annotations released with the dataset but only the text pairs that are selected for annotation. However, using lead author annotations instead of lexical level heuristics should increase paraphrase diversity in the released dataset beyond high lexical similarity pairs.

        \begin{table}[t]
        \centering \small
         \begin{tabular}{l r r} 
         \toprule
          \textbf{Split} & \textbf{\# (guest, host)-pairs} & \textbf{\# annotations} \\
         \midrule
            Train & 420 & 3896 \\ 
            Dev & 88 & 842 \\ 
            Test & 92 & 843 \\
         \midrule
            Total & 600 & 5,581 \\
         \bottomrule
        \end{tabular}%
        \caption{\label{fig:dataset-split} \textbf{Split of Dataset.} For each set, we show the number of text pairs and the total number of annotations.}
        \end{table}
            
        \textbf{Paraphrase Candidate Selection.} 
            We sample three datasets for annotation that differ in their estimated paraphrase distributions (based on the lead author annotations): 
            \textbf{BALANCED} is a set 100 text pairs sampled for equal representation of paraphrases and non-paraphrases.
            We annotate this dataset first with a high number of annotators per (guest, host)-pair, to decide on a crowd-worker allocation strategy that performs well for paraphrases as well as non-paraphrases.
            \textbf{RANDOM} is a uniform random sample of 100 text pairs. One main use of the dataset is to evaluate the quality of crowd-worker annotations on a random sample. 
            \textbf{PARA} is a set of 400 text pairs with an estimated 84\% of paraphrases designed to increase the variety of paraphrases in our dataset. Details on the sampling of the three datasets can be found in App.~ \ref{app:selection}.

\section{Annotation}\label{sec:annotation}

    We first describe the annotation task (\S \ref{sec:task}). Then, we discuss why the annotation task is difficult and a clear ground truth classification might not exist in many cases (\S \ref{sec:task-difficulty}). Therefore, we dynamically collect many judgments for text pairs with high disagreements (\S  \ref{sec:number-annotators}). 
    The annotation of utterance pairs takes place in two rounds with \texttt{Prolific} crowd-workers:
    (1) training crowd-workers (\S \ref{sec:annotator-training}) and
    (2) annotating paraphrases with trained crowd-workers (\S  \ref{sec:number-annotators} and \S \ref{sec:paraphrase-annotation}). 

   \begin{table}[t]
        \centering \small
         \begin{tabular}{l c c cc} 
         \toprule
         
          \textbf{Dataset} &  \multicolumn{2}{c}{\textbf{Guest}} &  \multicolumn{2}{c}{\textbf{Host}} \\ 
                      &  $\alpha$ &  $\frac{A\cap B}{A\cup B}$  &  $\alpha$ &  $\frac{A\cap B}{A\cup B}$ \\
        \midrule
            BALANCED & 0.42 & 0.51 & 0.48 & 0.63 \\  %
            RANDOM & 0.53 & 0.63 & 0.53 & 0.64 \\  %
            PARA & 0.43 & 0.50 & 0.50 & 0.64 \\  %
        \bottomrule
        \end{tabular}%
        \caption{\textbf{Agreement on highlights.} \label{tab:agreement-hls} 
            For pairs that at least two annotators classified a paraphrase, we display the average lexical overlap between the highlights (Jaccard Index displayed as $\frac{A\cap B}{A\cup B}$) and Krippendorff's unitizing $\alpha$ over all words for guest and host highlights, see \citet{krippendorff1995reliability}. 
        }
    \end{table}

\subsection{Annotation Task}

    \label{sec:task}

    Given a (guest, host) utterance pair, annotators (1) classify whether the host is paraphrasing any part of the guest's utterance and, if so, (2) highlight the paraphrase in the guest and host utterance. This results in data points like the one in Figure \ref{fig:figure-1}.
    Note that our setup differs from prior work, which usually involves classifying whether an entire text B is a paraphrase of an entire text A (e.g., \citealp{dolan-brockett-2005-automatically}). Instead, given texts A and B, our task is to determine whether there exists a selection of words from text B and text A, where the selection of text B is a paraphrase of the selection of text A. Our annotators are not only performing binary classification, but they also highlight the position of the paraphrase. To the best of our knowledge, we are the first to approach paraphrase detection in this way.
    Moreover, in contrast to previous work, the considered text pairs are usually longer than just one sentence and are contextualized dialog turns.

    \subsection{Plausible Label Variation}      
        \label{sec:task-difficulty}

    The task of annotating context-independent paraphrases is already difficult. Disagreements between human annotators are common \citep{dolan-brockett-2005-automatically, krishna-etal-2020-reformulating, kanerva_finish-paraphrase} --- even with extensive manuals for annotators \citep{kanerva_finish-paraphrase}. 
    In related semantic tasks like textual entailment,\footnote{
        Paraphrase classification has been repeatedly equated to (bi-)directional entailment classification \citep{dolan-brockett-2005-automatically, androutsopoulos2010survey}
        } 
    disagreements have been linked to plausible label variations inherent to the task %
    \citep{pavlick-kwiatkowski-2019-inherent, nie-etal-2020-learn, jiang-marneffe-2022-investigating}. 
    
    Our task setup adds further challenges: 
    First, instead of classifying full sentence pairs, annotators have to read relatively long texts and decide whether any portion of the text pair is a paraphrase. 
    Second, while in previous work annotators usually had to decide if two texts are generally approximately equivalent, they now need to identify paraphrases in a highly contextual setting with often incomplete information. 
    
   As a result, similar to the task of textual entailment, we expect classifying context-dependent paraphrases in dialog to not always have a clear ground truth. We display examples of plausible label variation in Table~\ref{tab:annotation-difficulties-examples}. 
    To handle label variation, common strategies are performing quality checks with annotators \citep{jiang-marneffe-2022-investigating} and recruiting a larger number of annotators for a single item \citep{nie-etal-2020-learn, sap-etal-2022-annotators}. We do both, see our approach in \S\ref{sec:annotator-training} and \S\ref{sec:number-annotators}.

               \begin{table}[t]
            \centering \footnotesize 
             \begin{tabular}{%
               p{70mm}} 
             \toprule
                \textbf{Shortened  Examples} \\ 
            \midrule
                \parbox{70mm}{
                    \textbf{G: }
                    \colorlet{orange100}{orange!100.0}\sethlcolor{orange100}\hl{we don't really know what went into their algorithm } {to make it turn out that way. } 
                     \\ 
                    \textbf{H: }
                    \colorlet{orange0}{orange!0}\sethlcolor{orange0}\hl{We're talking about algorithms, but } \colorlet{orange100}{orange!100.0}\sethlcolor{orange100}\hl{should we be talking about the humans who design the algorithms? } 
                    } \\
                    \midrule
                \parbox{70mm}{
                    \textbf{G: } \uline{In Harrison County.}
                     \\ 
                    \textbf{H: } \uline{In Harrison County.} Are you [...]
                    } \\

             \bottomrule
        \end{tabular}
        \caption{\textbf{Low Quality Annotations.} \label{tab:annotation-noise-example}
            We show human \sethlcolor{orange100}\hl{highlights} that can be considered wrong or noisy. When absent, we \uline{underline} the correct highlights. 
        }
        \end{table}  

     \begin{table*}[t]
        \centering \small
         \begin{tabular}{l  c c c c  r c c} 
         \toprule
            & \multicolumn{4}{c}{\textbf{Classification}} & \multicolumn{3}{c}{\textbf{Highlighting}}\\
            \textbf{Model}  & Extract~$\downarrow$ &  F1~$\uparrow$ &  Prec~$\uparrow$ & \multicolumn{1}{c}{Rec~$\uparrow$} & {Extract~$\downarrow$} & Jacc Guest~$\uparrow$ & Jacc Host~$\uparrow$  \\ 
        \midrule
                llama 2 7B 
                    & \uline{1\%} & {0.66} & 0.49 & \textbf{0.98}  
                    & 59\% & 0.34 & 0.44  \\
                vicuna 7B 
                    & \textbf{1\%} & 0.29 & 0.67 & 0.19  
                    & 32\% & 0.30 & 0.46 \\
                Mistral 7B Instruct v0.2  
                    & 3\% & 0.62 & 0.66 & 0.58 
                    & 66\% & 0.40 & {0.51}  \\ 
                openchat 3.5
                    & \textbf{0\%} & {0.66} & {0.76} & {0.58} 
                    & 64\% & 0.46 & {0.50}  \\
                gemma 7B
                    & \uline{1\%} & {0.64} & {0.66} & {0.63}  
                    & 48\% & 0.24 & {0.51}  \\
                Mixtral 8x7B Instruct  v0.1 
                    & \textbf{0\%} & \uline{0.74} & 0.73 & 0.74 
                    & 65\% & {0.35} & 0.52  \\
                Llama 2 70B %
                    & \textbf{0\%} & 0.66 &  0.72 & 0.61  
                    & 71\% & {0.29} & {0.56}  \\                     
                GPT-4 
                    & \textbf{0\%} & \textbf{0.81} & \uline{0.78} & \uline{0.84}  
                    & \uline{17\%} & \textbf{0.67} & \textbf{0.71}  \\
                \midrule
                DeBERTa v3 large AGGREGATED
                & \textbf{-} & {0.73}  & {0.67}  & {0.81}  & \textbf{-} & \uline{0.52} & \uline{0.66} \\

                DeBERTa v3 large ALL & \textbf{-} & {0.66}  & \textbf{0.82}  & {0.56}  & \textbf{-} & {0.45} & {0.64} \\
        \bottomrule
        \end{tabular}%
        \caption{\textbf{Modeling Results.} \label{tab:model-results} 
            We \textbf{boldface} the best and \uline{underline} the second best performance. We display the extraction error of predictions from generative models and, for classification, the F1, precision and recall score as well as, for highlights, the Jaccard Index for the guest and host utterances. Higher values are better ($\uparrow$) except for extraction errors ($\downarrow$). 
            \texttt{GPT-4} is the best classification model, while, overall, \texttt{DeBERTa} is the best highlight model as it does not lead to any extraction errors.
        }
    \end{table*}

    \subsection{Annotator Training} \label{sec:annotator-training}
    
    When annotating paraphrases, the instructions for annotators are often short, do not explain challenges and rely on annotator intuitions \citep{dolan-brockett-2005-automatically,lan-etal-2017-continuously}.\footnote{
        For example, instructions are to rate if two sentences ``mean the same thing'' \citep{dolan-brockett-2005-automatically} or are ``semantically equivalent''   \citep{lan-etal-2017-continuously}.
        }
    In contrast, \citet{kanerva_finish-paraphrase} recently used an elaborate 17-page manual. However, they relied on only 6 expert annotators that might not be able to represent the full complexity of the task (\S\ref{sec:task-difficulty}). We aim for a trade-off between short intuition-based and long complex instructions that facilitates recruitment of a larger number of annotators: an accessible example-centric, hands-on 15-minute training of annotators that teaches our operationalization of context-dependent paraphrases (\S \ref{sec:paraphrase-operationalization}). We provide (1) a short paraphrase definition, (2) examples of context-dependent paraphrases showing clear and approximate equivalence (c.f.~Table~\ref{tab:CP-examples}), (3) examples of common difficulties with paraphrase classification in dialog (c.f.~Table \ref{tab:NCP-examples} and \S \ref{sec:paraphrase-operationalization}), and use (4) a hands-on approach where annotators have to already classify and highlight paraphrases  after receiving instructions. Only once they make the right choice on what is (Table \ref{tab:CP-examples}) and is not a paraphrase~(Table \ref{tab:NCP-examples}) and highlight the correct spans they are shown the next set of instructions. Only annotators that undergo the full training and pass two comprehension and two attention checks are part of our released dataset. Overall, 49\% of the annotators who finished the training passed it. 
    See App.~\ref{app:paraphrase-annotations} for the instructions and further details. %

    \subsection{Annotator Allocation} \label{sec:number-annotators}

      To the best of our knowledge, text pairs in paraphrase datasets receive a fixed number of 1, up to a maximum of 5 annotations \citep{kanerva_finish-paraphrase,zhang-etal-2019-paws,lan-etal-2017-continuously,dolan-brockett-2005-automatically}. 
      However, this might not be enough to represent the inherent plausible variation to the task (\S\ref{sec:task-difficulty}). 
        We have each pair in BAL\-ANCED annotated by 20--21 trained annotators to simulate different annotator allocation strategies (App.~\ref{app:annotator-allocation}). Then, for RANDOM and PARA, we use a dynamic allocation strategy: Each pair receives at least 3 annotations. We dynamically collect more annotations, up to 15, on pairs with high disagreement (i.e., entropy $>$ 0.8). 
        Overall, this results in an average of 9 annotations per text pair across our released dataset.

\subsection{Results} \label{sec:paraphrase-annotation} \label{sec:annotation-result}

    We discuss annotations results (tables \ref{tab:agreement}, \ref{tab:label-dist}, \ref{tab:agreement-hls}) on our  datasets BALANCED, RANDOM and PARA.
    
        \textbf{Classification agreement as an indicator of variation.} 
        Agreement for classification is relatively low (Table \ref{tab:agreement}). %
        We inspect a sample of 100 annotations %
        on the RANDOM set and manually assess annotation quality.  90\% of the annotations can be said to be at least plausible (see Table~\ref{tab:annotation-noise-example} for low quality and Table~\ref{tab:annotation-difficulties-examples} for plausible variation examples), which is in line with the fact that we only use high quality annotators (\S \ref{sec:annotator-training}). Further, we manually analyze the 42 annotations of ten randomly sampled annotators: Nine annotators consistently provide high quality annotations, while the other annotator chooses “not a paraphrase” a few times too often (see Appendix \ref{app:intra-annotator-annotations} for details).
        As a result, we assume that most disagreements are due to the inherent plausible label variation of the task (\S \ref{sec:task-difficulty}).

        \textbf{Higher agreement on paraphrase position.} Krippendorff's unitizing $\alpha$ on the highlights is higher than in other areas\footnote{
            E.g., 0.41 for hate speech \citep{carton-etal-2018-extractive} or 0.35 for sentiment analysis \citep{sullivan-etal-2022-explaining}. Because of the different tasks these values are not exactly comparable.
        } 
        (see Table \ref{tab:agreement-hls}). We also calculate the ``Intersection-over-union'' between the highlighted words (i.e., Jaccard Index), a common and interpretable evaluation measure for annotator highlights \citep{herrewijnen2024human,mendez-guzman-etal-2022-rafola, Mathew_Saha_Yimam_Biemann_Goyal_Mukherjee_2021, malik-etal-2021-ildc}.
        It seems that while annotations vary on whether there is a paraphrase or not, they agree  frequently on the position of the possible paraphrase. On average, at least 50\% of the highlighted words are the same between annotations.\footnote{
            100\% overlap in highlighting is uncommon. \citet{deyoung-etal-2020-eraser} consider two highlights a match if Jaccard is greater than 50\%.
        }  
        Agreement is higher on the host utterance, because on average the host utterance is shorter than the guest utterance (33 $<$ 85 words). 

        \textbf{Label variation is highest for paraphrases.}
        Between the datasets, classification agreement is lowest for PARA. This is what we expected since it has the largest portion of ``hard'' non-repetition paraphrases (see App. \ref{app:selection}). Krippendorff's $\alpha$ is lower for the RANDOM than the BALANCED set, even though we expected the RANDOM set to include easier decisions for annotators (RANDOM includes more unrelated non-paraphrases, see App. \ref{app:selection}). As the other agreement heuristic is relatively high on RANDOM, the lower $\alpha$ values could be a result of Krippendorff's measure being sensitive to imbalanced label distributions \citep{riezler2022validity}, see also Table  \ref{tab:label-dist} displaying the imbalanced distribution for RANDOM.

     \begin{table}[t]
            \centering \footnotesize 
             \begin{tabular}{
               p{1mm} p{1mm} p{1mm}  p{55mm}} 
             \toprule
                & \multicolumn{2}{ p{3mm}}{\textbf{Preds}}
                &  \textbf{Shortened  Examples} \\
                T & G & D\vspace*{-0.4\baselineskip}  \\ 
            \midrule
                \xmark & \xmark & \uline{\cmark}\vspace*{-0.4\baselineskip} &
                    \parbox{56mm}{
                    \textbf{G: }
                        He was \uline{the most famous guy} in the world of sports... 
                     \\ 
                    \textbf{H: }
                    \uline{The most famous Italian...}
                } \\ \midrule
               \cmark & \xmark & \uline{\cmark}\vspace*{-0.4\baselineskip} & 

                    \parbox{56mm}{
                    \textbf{G: } 
                    {A lot of them were the Bay Area influx that came up and \uline{bought homes to flip}. \sethlcolor{orange100}\hl{You know what flipping is}, right?}
                              \\ 
                    \textbf{H: } Mm-hmm. \hluline{orange100}{Buying a house, improving} \hluline{orange100}{ it, selling it out of profit}.
                    }
                    \\
            \bottomrule
        \end{tabular}
        \caption{
            \textbf{Model Errors.}
            \label{tab:deduplicate-error-analysis} 
            We show examples of prediction errors made by \texttt{DeBERTa} (D) and \texttt{GPT-4} (G). We display model predictions (D/G) for paraphrases (\cmark) and non-paraphrases (\xmark) and compare it to the crowd-majority (T). If one model  predicted a paraphrase the corresponding text spans are \uline{underlined}. For comparison, we also display the crowd majority \sethlcolor{orange100}\hl{highlights}. 
        }
        \end{table}

\section{Modeling}

    In Table~\ref{fig:dataset-split}, we do a random 70, 15, 15 split of our 5,581 annotations, along the 600 unique pairs. 
    
    \textbf{Token Classifier.} 
    Similar to \citet{wang-etal-2022-paratag}, we fine-tune a large \texttt{DeBERTa}  model\footnote{\href{https://huggingface.co/microsoft/deberta-v3-large}{\texttt{microsoft/deberta-v3-large}}} \citep{he2020deberta} on token classification to highlight the paraphrase positions (for hyperparameters, see App.~\ref{app:token-classification}). We train two models: using all 3,896 training annotations (``ALL'' in Table \ref{tab:model-results}) and using the majority aggregated training annotations over the 420 unique (guest, host) training pairs (``AGGREGATED'' in Table \ref{tab:model-results}). We consider a model to have predicted a paraphrase for a pair if at least one token is highlighted with softmax probability $\geq0.5$ in both texts. For each model, we average performances over three seeds.

    \textbf{In-Context Learning.}
    We further prompt the following generative models (see URLs in App. \ref{app:icl-models}) to both classify and highlight the position of paraphrases: \texttt{Llama 2 7B} and \texttt{70B} \citep{touvron2023llama}, 
    \texttt{Vicuna 7B} %
    \citep{vicuna},
    \texttt{Mistral 7B Instruct v0.2} %
    \citep{jiang2023mistral}, 
    \texttt{Openchat 3.5} %
    \citep{wang2023openchat}, 
    \texttt{Gemma 7B} %
    \citep{team2024gemma},
    \texttt{Mixtral 8x7B Instruct v0.1} %
    \citep{jiang2024mixtral} and 
    \texttt{GPT-4}\footnote{API calls where performed using the ``gpt-4'' model id in March 2024.} \citep{achiam2023gpt}. 
    We design the prompt to be as close as possible to the annotator training using a few-shot setup  \citep{FEW-SHOT_NEURIPS2020_BROWN, pmlr-v139-zhao21c} with all 8 examples shown during annotator training. We also provide explanations in the prompt \citep{chain-of-thought, NLI-explanations} and use self-consistency by prompting the models 10 (\texttt{GPT-4} and \texttt{Llama~70B}:~3) times \citep{wang2022self}. For the prompt and further hyperparameter settings see App. \ref{app:prompting}.

    \begin{table}[t]
            \centering \footnotesize 
             \begin{tabular}{
               p{0.91\columnwidth}} 
             \toprule
               \textbf{Shortened  Example} \\
            \midrule
               
                    \colorlet{orange66}{orange!66.67}
                    \colorlet{orange33}{orange!33.33}
                    \textbf{G: }
                        ... then \sethlcolor{orange33}\hl{he goes on and}\hluline{orange33}{ references }\hluline{orange66}{ and} \hluline{orange100}{ \textbf{makes mention}}\hluline{orange100}{\textbf{of Rudy Giuliani three times} in this} \hluline{orange100}{ conversation} 
                     \\ 
                    \textbf{H: }
                        And  \colorlet{orange66}{orange!66.67}\hluline{orange66}{\textbf{Rudy Giuliani}} \colorlet{orange33}{orange!33.33}\sethlcolor{orange33}\hl{was a private lawyer not a government official,} \sethlcolor{orange66}\hl{so} \sethlcolor{orange100}\hl{why is} 
                        \hluline{orange100}{\textbf{he coming up}} \sethlcolor{orange100}\hl{\textbf{ so}} \hluline{orange100}{\textbf{much in}} \sethlcolor{orange100}\hl{\textbf{this}} \hluline{orange100}{\textbf{conversation}} \sethlcolor{orange66}\hl{between two world leaders?}
                 \\
            \bottomrule
        \end{tabular}
        \caption{
            \textbf{Highlighting Differences.}\label{tab:deduplicate-hl-error-analysis} 
            We show examples of highlights made by \uline{\texttt{DeBERTa}},  \textbf{\texttt{GPT-4}} and {\sethlcolor{orange100}\hl{human highlights}}.   
            {\sethlcolor{orange33}\hl{Lower intensity}} means less human annotators selected the word. 
            While \texttt{GPT-4} struggles with providing highlights at all (c.f. extraction error in Table~\ref{tab:model-results}), \texttt{DeBERTa} highlights tend to be too sparse (just ``Rudy Giuliani'', ``coming'' and ``conversation'' in the host utterance).
            Here, we highlight words, when the softmax probability is  $>0.44$\tablefootnote{We tried a few different thresholds $>0.40$ with $0.44$ getting the biggest gain in the Jaccard Index on the test set.} instead of $\geq0.5$. On the complete test set, this also increases the mean Jaccard Index (by $0.06$/$0.01$ for guest/host compared to Table~\ref{tab:model-results}).
        }
        \end{table}

   \textbf{Results.}
   For evaluation, we consider a pair to contain a paraphrase if it has been classified by a majority of crowd-workers and a word to be part of the paraphrase if it has been highlighted by a majority of crowd-workers. We leave soft-evaluation approaches to future work \citep{uma2021learning}, among others because of challenges in extracting label distributions for in-context learning in a straight-forward way \citep{hu-levy-2023-prompting, lee-etal-2023-large}. 
   See Table \ref{tab:model-results} for test set performances. Performances for the token classifier are the mean over three seeds. Performances for the generative models is the majority vote for the 3--10 self-consistency calls. We display the F1 score for classification and, as before (\S\ref{sec:annotation-result}), Intersection-Over-Union of the highlighted words for guest and host utterance highlights (Jaccard-Indices), see, for example, \citet{deyoung-etal-2020-eraser}. For in-context learning, we also display how often we could not extract the highlights or classifications from  model responses. 
   Note that the test set contains 93 elements, so differences between models might appear bigger than they are. 
   
   Overall, \texttt{GPT-4} and \texttt{Mixtral 8x7B} achieve the best results in paraphrase classification. In highlighting, our \texttt{DeBERTa} token classifiers and \texttt{GPT-4} achieve the best overlap with human annotations. 
   However, due to problems with extracting highlights from  model responses (e.g., hallucinations, see App.~\ref{app:error-analysis}), our fine-tuned \texttt{DeBERTa} token classifiers are probably the best choice to extract the position of paraphrases. While the \texttt{DeBERTa AGGREGATED} model achieves higher F1 scores, the \texttt{DeBERTa ALL} model has the highest precision out of all models.
   We provide our best-performing \texttt{DeBERTa AGGREGATED} model (model with seed $202$ and F1 score of $0.76$) on the Hugging Face Hub\footnote{\url{https://huggingface.co/AnnaWegmann/Highlight-Paraphrases-in-Dialog}} and use it in the following error analysis.

   \textbf{Error Analysis.}
   We consider the best-performing classification and highlighting models for error analysis, i.e., \texttt{GPT-4} and \texttt{DeBERTa AGGREGATED}.
    We manually analyze a sample of misclassifications, for examples  see Table \ref{tab:deduplicate-error-analysis}. 
        Overall, the classification quality is better for \texttt{GPT-4}. 
        The \texttt{DeBERTa} classifier finds more paraphrases (note that \texttt{DeBERTa AGGREGATED} for seed $202$ has a recall of $0.86$) but also predicts more false positives than \texttt{GPT-4}. 
   For both models, the items with incorrect predictions also show higher human disagreement. The average entropy for human classifications
   is lower for the correct ($0.45$ for \texttt{DeBERTa}, $0.45$ for \texttt{GPT-4}) than for the incorrect model predictions ($0.59$ for \texttt{DeBERTa}, $0.67$ for \texttt{GPT-4}).
   \texttt{DeBERTa} highlights shorter spans of text (on average $6.6$/$6.2$, compared to $16.7$/$10.9$ for \texttt{GPT-4} for guest/host respectively), while \texttt{GPT-4} usually highlights complete (sub-)sentences. %
   \texttt{GPT-4} highlights are largely of good quality, however they often can not be extracted (see App.~\ref{app:error-analysis}). The \texttt{DeBERTa} highlights can seem ``chopped up'' and missing key information (e.g., the original host highlights in  Table \ref{tab:deduplicate-hl-error-analysis} are just ``Rudy Giuliani'', ``coming'' and ``conversation''). 
   We recommend performing a classification of an utterance pairs as a paraphrase when there exist softmax probabilities $\geq 0.5$ for both guest and host utterance, but then selecting the highlights also based on softmax probabilities lower than $0.5$. Alternatively, the best \texttt{DeBERTa ALL} model\footnote{\url{https://huggingface.co/AnnaWegmann/Highlight-Paraphrases-in-Dialog-ALL}} provides fewer but seemingly more consistent highlights (see Appendix~\ref{app:error-analysis}). One possible reason for this could be that \texttt{DeBERTa ALL} was trained on individual highlights provided by single annotators, rather than on aggregated highlights.

\section{Conclusion}

    A majority of work on paraphrases in NLP has looked at the semantic equivalence of sentence pairs in context-independent settings. However, the human dialog setting is highly contextual and typical  methods fall short.
    We provide an operationalization of context-dependent paraphrases and an up-scalable hands-on training for annotators. We demonstrate the annotation approach by providing 5,581 annotations on a set of 600 turn pairs from news interviews. Next to paraphrase classifications, we also provide annotations for paraphrase positions in  utterances.
    In-context learning and token classification both show promising results on our dataset.
    With this work, we contribute to the automatic detection of paraphrases in dialog.  We hope that this will benefit both NLP researchers in the creation of LLMs and social science researchers in analyzing paraphrasing in human-to-human or human-to-computer dialogues on a larger scale.

\section*{Limitations}

    Even though the number of our unique text pairs is relatively small, we release a  high number of high quality annotations per text pair  (5,581 annotations on 600 text pairs). Releasing more annotations on fewer ``items'' (here: text pairs), has increasingly been more common in NLP \citep{nie-etal-2020-learn,sap-etal-2022-annotators}.
    Further, big datasets become less necessary with better generative models: 
    Using only eight paraphrases pairs in our prompt already led to promising results.
    We further use the full 3,896 annotations from the training set to train a token classifier showing competitive results with the open generative models. However, the token classifier and other potential fine-tuning approaches would probably profit from a bigger dataset.

    Even though our dataset of news interviews showed frequent, different and diverse occurrences of paraphrasing, it might not be representative of paraphrasing behavior in conversations across different contexts and social groups. In the future, we aim to expand our dataset with further out-of-domain items.
    
    Our data creation process was not aimed at scalability. While our developed annotator training procedure can easily be scaled to a larger group of crowd-workers, we manually selected text pairs for annotation. Future work could scale this by skipping manual selection and accepting a more imbalanced dataset or using our trained classifiers as a heuristic to identify likely paraphrases. 

    Even though we carefully prepared the annotator training and took several steps to ensure high-quality annotations, there remain several choices that were out of our scope to experiment with, but might have improved quality even more. For example, experimenting with different visualizations of paraphrase highlighting, text fonts, giving annotators an option to add confidence scores for classifications and so on. 

    We only use one prompt that is as close as possible to the instructions the human annotators receive. We use the same prompt with the exact same formatting for all different generative LLMs. However, experimenting with different prompts might improve performance \citep{weng2023prompt} and some models might benefit from certain formatting  %
    or phrasing. We leave in-depth testing of prompts to future work.
    Further, it might be possible to improve the performance of our \texttt{DeBERTa} model, through providing contextual information (like speaker names and interview summary). Currently, these are only provided to the generative models.

    In this work we collect a high number of human annotations per item and highlight the plausible label variation in our dataset. However, we use hard instead of soft-evaluation approaches \citep{uma2021learning} for the computational models. We do this because, among others, extracting label distributions for in-context learning is challenging \citep{hu-levy-2023-prompting, lee-etal-2023-large}. We leave the development of a soft evaluation approach to future work but want to highlight the potential of our dataset here:
    The high number of annotations per item enables the modeling of classifications and text highlights as distributions, similar to \newcite{zhang-de-marneffe-2021-identifying}. Further, our dataset provides anonymized unique ids for all annotators and enables modeling of different perspectives, e.g., with similar methods to \newcite{sachdeva-etal-2022-measuring} and \newcite{deng-etal-2023-annotate}.

    We do not differentiate between different communicative functions, intentions or strategies that affect the presence of paraphrases in a dialog. This is relevant as paraphrases might, for example, be a more conscious choice by interviewers \citep{clayman2002news} or a more unconscious occurrence similar to the linguistic {alignment} of the references for discussed objects \citep{xu-reitter-2015-evaluation,garrod1987alignment}. %
    With this work, we hope to provide an outline of the general class of context-dependent paraphrases in dialog that lays the groundwork for further, fine-grained distinctions.

\section*{Ethical Considerations}

    We hope that the ethical concerns of reusing a public dataset \citep{zhu-etal-2021-mediasum} are minimal. Especially, since the CNN and NPR interviews are between public figures and were broadcast publicly, with consent, on national radio and TV.

    Our dataset might not be representative of English paraphrasing behavior in dialogs across different social groups and contexts as it is taken from U.S. news interviews with public figures from two broadcasters. %
    We caution against using our models without validation on out-of-domain data. 

    We performed several studies with U.S.-based crowd-workers as part of this work. We payed participants a median of $\approx11.41\$$/h which is above federal minimum wage. Crowd-workers consented to the release of their annotations. We do not release identifying ids of crowd-workers.
    
    We confirm to have read and that we abide by the ACL Code of Ethics.
    Beside the mentioned ethical considerations, we do not foresee immediate risks of our work.

\section*{Acknowledgements} 
    We thank the anonymous ARR reviewers for their constructive comments. 
    Further, we thank the NLP Group at Utrecht University and, specifically, Elize Herrewijnen, Massimo Poesio, %
    Kees van Deemter, Yupei Du, Qixiang Fang, Melody Sepahpour-Fard, Shane Kaszefski Yaschuk, Pablo Mosteiro, and Albert Gatt, for, among others, feedback on writing and presentation, discussions on annotator disagreement and testing multiple iterations of our annotation scheme. We thank Charlotte Vaaßen, Martin Wegmann and Hella Winkler for feedback on our annotation scheme. We thank Barbara Bziuk for feedback on presentation.
    This research was supported by the “Digital Society - The Informed Citizen” research programme, which is (partly) financed by the Dutch Research Council (NWO), project 410.19.007. %

\bibliography{anthology_p1, anthology_p2, custom}

\clearpage
\newpage

\appendix 

\section{Context-Dependent Paraphrases in Dialog}

    \paragraph{Should one include repetitions?} Repetitions have been typically included in paraphrase taxonomies \citep{bhagat-hovy-2013-squibs, zhou2022paraphrase} even though, e.g., \citet{kanerva_finish-paraphrase} asked annotators to exclude such pairs as they considered them uninteresting paraphrases. However, distinguishing repetitions from paraphrases turns out to be especially hard in dialog: speakers tend to leave words out when they repeat and adapt the pronouns to match their perspective (e.g., I -> you). We therefore include repetitions in our definition of context-dependent paraphrases. In fact, those mainly make up the ``Clear Contextual Equivalence'' Paraphrases (see Table \ref{tab:CP-examples}).

    \begin{table}[t]
    \centering \small
     \begin{tabular}{l c r c c c c} 
     \toprule
      & \multicolumn{2}{c}{Preprocessed} & \multicolumn{2}{c}{Sampled} & \multicolumn{2}{c}{Released} \\
      &  \# i & \# gh &  \# i & \# gh &  \# i & \# gh \\ 
     \midrule
      all &  
                34419 & 148522  %
                    & 1304 & 4450 & 480 & 600 \\ 
        \ NPR & 11506 & 49065   %
                    & 423 & 1550 & 167 & 218 \\ 
        \ CNN & 22913 &  99457  %
                    & 881 & 2900 & 313 & 382 \\
     \bottomrule
    \end{tabular}%
    \caption{\label{fig:final-dataset-statistics} \textbf{Dataset Statistics.} Number of interviews (\#i) and (guest, host)-pairs (\# gh) respectively after preprocessing (\S \ref{sec:preprocessing}), random sampling (\S \ref{sec:candidate-selection}) and the selection of paraphrase candidates for annotation (\S \ref{sec:candidate-selection}).}
    \end{table}

\section{Dataset} \label{sec:interface}

        \paragraph{Topic of the Dataset.} The topics of the CNN and NPR news interviews \citep{zhu-etal-2021-mediasum} are mostly centered around U.S. politics (e.g., presidential or local elections, 9/11, foreign policy in the middle east), sports (e.g., baseball, football), domestic natural disasters or crimes and popular culture (e.g., interviews with book authors). %

        \paragraph{Utterance Pair IDs.} We use unique IDs for utterance pairs. For example, for NPR-4-2, ``NPR-4'' is the ID used for interviews\footnote{In this case referring to \url{https://www.npr.org/templates/story/story.php?storyId=16778438}} as done in \citet{zhu-etal-2021-mediasum}, 
        ``2'' is the position of the start of the guest utterance in the utterance list as separated into turns by \citet{zhu-etal-2021-mediasum}, in this case ``Thank you.''.

    \subsection{Preprocessing}

        We give details on the three preprocessing steps (see \S \ref{sec:preprocessing}).

        \textbf{1. Filtering for 2-person interviews.} \label{app:data-2person-filter}
        We filter 49,420 NPR and 414,176 CNN interviews from \citet{zhu-etal-2021-mediasum} for 2-person interviews only.
        This can be challenging: In the speaker list, authors sometimes have non-unique identifiers (e.g., `STEVE PROFFITT', `PROFFITT' or `S. PROFFITT' refer to the same speaker). If one author identifier string is contained in the other we assume them to be the same speaker.\footnote{
            There might be other cases where different string identifiers in the dataset refer to the same speaker although they are not substrings of the other (e.g.,  `S. PROFFITT' and `STEVE PROFFITT'). For a randomly sampled selection of 44 interviews that were identified as more than 2 person interviews, 12 contained errors in the matching. %
            2/12 were the result of typos and 10/12 were the result of additions to the name like ``(voice-over)'' or ``(on camera)''. %
        } 
        We generally assume the first speaker  to be the host. We remove 538 NPR and 1,917 CNN interviews because the identifier of the second speaker includes the keywords ``host'' or ``anchor'' --- thus contradicting our assumption.
        This leaves 14,000 NPR and 50,301 CNN 2-person interviews. %

        \textbf{2. Removing first and last turns of an interview.}
            The first turns in our 2-person interviews are usually (reactions to) welcoming addresses and acknowledgments by host and guest\footnote{For example,
            ``I'm Farai Chideya.'' 
            ``Welcome.''
            ``Thank you.''
            }, while the last often contain goodbyes or acknowledgments\footnote{ 
            For example the last 3 turns in the considered NPR-4 interview:
            ``Well, Dr. Hader. Thanks for the information.'', ``Well, thank you for helping share that information [...]'', ``Well, thanks again. Dr. Shannon Hader [...]''}. 
            We remove the first two and the last two (guest, host)-pairs. 
            This step removes 2,409 NPR and 26,419 CNN interviews because they are fewer than 5-turns long.  For the remaining  interviews, this removes 34,773 NPR and 71,646 CNN (guest, host)-pairs.
        
        \textbf{3. Removing short and long  utterances.}        
        We further remove short guest utterances of 1--2 words as they leave not much to paraphrase.\footnote{
        We manually looked at a random sample of $0.3\%\approx48$ such pairs.  The 1-2 token guest utterances are mostly (40/48) assertions of reception by the guest (e.g., ``Yes.'', ``Exactly. Exactly.'', ``That's right''). Some are signals of protest   (4/48) (e.g., ``Hey, man.'', ``Yes, but...'', ``Hold on.''). None of them were reproduced by the host in the next turn.
        } 
        3,540 NPR and 12,675 CNN pairs are removed like this.
        We also remove pairs where the host utterance consists of only 1--2 words.\footnote{
        We manually looked at a random sample of $0.3\%\approx37$ such pairs. The 1--2 tokens host utterances are mostly (28/37) assertions of reception by the host (e.g., ``Yeah.'', ``Yes.'', ``Sure.'', ``Right.'', ``Right. Right.'',  ``Ah, okay.''). Some are requests for elaboration (5/37) (e.g., ``How so?'', ``Like?'', ``Four?'') or reactions (3/37) (e.g., ``Wow!'', ``Oh, interesting.''). %
        Only one example ``Four?'' was reproducing content in the form of a repetition. 
        }. %
        2,940 NPR and 11,389 CNN pairs 
        are removed like this. 
        We also remove pairs  where guest or host utterance consist of more than 200 words.\footnote{ %
        200 is the practical limit for the number of words for the chosen type of question (i.e., `Highlight'' Question) in the used survey hosting platform (i.e., \texttt{Qualtrics}). It also limits annotation time per question.}
        Overall, this leaves 148,522 (guest, host)-pairs in 34,419 interviews for potential annotation, see Table \ref{fig:final-dataset-statistics}. 

    \begin{figure}[t]
        \centering
        \includegraphics[width=\columnwidth]{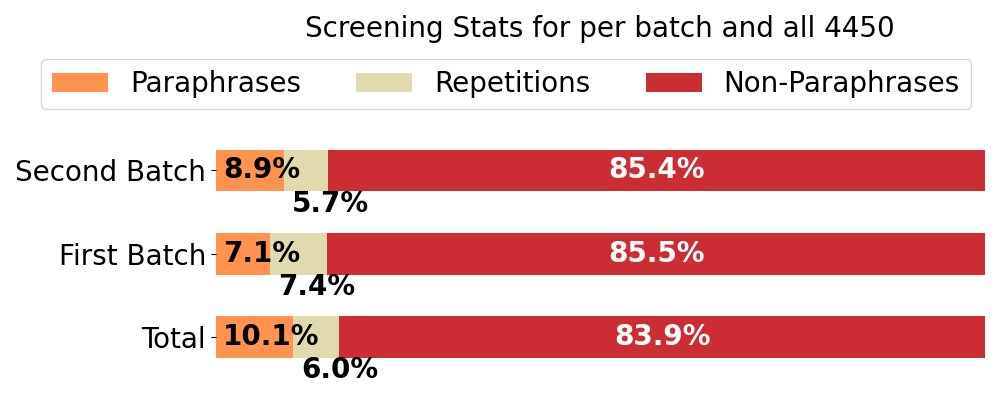}
        \caption{\textbf{Label distribution after first author annotations performed in two batches.} First author label classification was performed in two batches. The first batch consists of 750 text pairs, the second of 3,700.}
        \label{fig:dist-pc-batches}
    \end{figure}

    \subsection{First Author Annotations} \label{app:paraphrase-candidates}

        We provide more details on the first author annotations for selecting paraphrase candidates (\S\ref{sec:annotation-sets}).

        \textbf{Deciding on first author annotations.} Since the share of paraphrases in randomly sampled (guest, host)-pairs  was only at around 5-15\% in initial pilots with lab members, similar to previous work, we opted to do a pre-selection of text pairs before proceeding with the more resource-intensive paraphrase annotation (c.f.~\S \ref{sec:paraphrase-annotation} and App. \ref{app:paraphrase-annotations}). However, commonly used automatic heuristics were not suitable for the highly contextual discourse setting (c.f.~\S \ref{sec:candidate-selection}). Instead, we experimented with discarding obvious ``non-paraphrases'' through crowd-sourced annotations and compared it to manual annotations by the lead author, ultimately deciding on using lead author annotations. One of the reasons was that discarding obvious ``non-paraphrases'' was more resource intensive and difficult for crowd-workers than expected, making the resources needed for discarding non-paraphrases too close to annotating paraphrases themselves -- which defeats the purpose of doing a pre-selection in the first place.

        \begin{table}[t]
        \centering \small
         \begin{tabular}{l r} %
         \toprule
            \textbf{Paraphrase}       & 88 \\ 
            \midrule
            \ \ High Lexical Similarity      & 59 \\ 
            \ \ \ \ Repetition            & 45 \\
            \ \ Perspective-Shift     & 10 \\ 
            \ \ Directional           & 17 \\ 
            \ \ Difficult Decision & 16 \\
            \midrule
            \textbf{Non-Paraphrase } & 519 \\ 
            \midrule
            \ \ High Lexical Similarity     & > 18 \\ 
            \ \ Partial              & > 24 \\
            \ \ Unrelated            & > 103 \\
            \ \ Topically Related    & > 83 \\
            \ \ Conclusion           & 46 \\ 
            \midrule
            \textbf{Ambiguous}  &  18 \\
        \midrule
            \textbf{Missing Context} & 125 \\ 
        \bottomrule
        \end{tabular}%
        \caption{\textbf{Statistics Labels First Batch.} For 750 manually reviewed pairs, we also labeled several other categories. We found 88 paraphrases, 519 non-paraphrases, 18 ambiguous cases and 125 where the missing context impeded a definite decision. Note that we tried to not assign ambiguous if we were leaning to one category over another. 
        Other categorizations include: 
        ``perspective-shift'' (the perspective shifts between guest and host, e.g., ``you'' -> ``I''), 
        ``directional'' (guest or host utterance is entailed from or subsumed in the other), %
        ``partial'' (a subsection could be understood as a paraphrase, but the overall larger section is clearly not a paraphrase),
        ``related'' (two utterances are closely related but no paraphrases), ``conclusion'' (host draws a conclusion or adds an interpretation that goes beyond a paraphrase).  Some labels were only added in the last 200 annotations and therefore include the ``>'' indication. 
        }
        \label{tab:pc-statistics}

        \end{table}
        
        \begin{table}[h!]
            \centering \small
             \begin{tabular}{l c} 
             \toprule
             
              Dataset & Overlap Lead and Crowd\\
            \midrule
                BALANCED & 0.72  \\  
                RANDOM & 0.89  \\ 
                PARA & 0.72 \\  
            \bottomrule
            \end{tabular}
            \caption{\label{app:lead-crowd}\textbf{Lead vs. Crowd Classifications.} We display the average overlap between the lead author's classifications and the majority vote of the crowd. The overlap is the highest on the RANDOM set. Probably because we keep all obvious non-paraphrases for classification and the annotators face less ambiguous (guest, host)-pairs to classify.}
        \end{table}  
        
        \textbf{Changing lead author annotations from discarding obvious non-paraphrases to keeping interesting paraphrases.} On an initial set of 750 random (guest, host)-pairs, we remained with the initial idea of discarding  obvious non-paraphrase pairs. 
        However, due to a resulting high share of uninteresting or improbable paraphrase pairs, %
        we opted to classify paraphrases vs. non-paraphrases instead of possible paraphrases vs. obvious non-paraphrases. The lead author re-annotated the initial set of 750 paraphrase candidates and annotated 4450 additional (guest, host)-pairs for paraphrase vs. non-paraphrase. In the first batch, the lead author additionally labeled a variety of different paraphrase types/difficulties (e.g., high lexical similarity, missing context, unrelated), see also Table \ref{tab:pc-statistics}, in the second batch this was restricted to repetition paraphrase, paraphrase and non-paraphrase. 
        The distribution of these three categories is displayed in Figure~\ref{fig:dist-pc-batches}. %

        \textbf{Relation to with Crowd Majority Annotations.} \label{app:lead-crowd}
        We display the overlap between the lead author's paraphrase classifications and the released classifications of the crowd majority in Table \ref{app:lead-crowd}.

    \begin{table}[t]
        \centering \small
         \begin{tabular}{l | c | c } 
         \toprule
           Type (guest, host)-pair &  \#  & acc. \\ 
        \midrule
            \textbf{Paraphrase}      & 46 & 0.80 \\  %
            \midrule
            \ \ High Lexical Similarity & 24  & 0.92 \\ 
            \ \ \ \ Repetition & 16  & 0.88 \\  %
            \ \ Context-Dependent  & & \\ %
            \ \ \ \ Perspective-Shift     & 10  & 0.90 \\
            \ \ \ \ Directional & 12  & 0.67 \\ 
            \ \ Other Difficult Cases & 12  & 0.58 \\  %
            \midrule
            \textbf{Non-Paraphrase } & 54  & 0.81 \\
            \midrule
            \ \ Unrelated utterances & 13  & 1.00 \\
            \ \ More Difficult & 41  & 0.76 \\
            \ \ \ \ Topically related           & 24  & 0.67 \\
            \ \ \ \ High Lexical Similarity     & 11  & 0.64 \\
            \ \ \ \ Partial              & 10  & 0.80 \\
            \ \ \ \ Conclusion            & 11  & 0.55 \\ 
        \bottomrule
        \end{tabular}%
        \caption{\textbf{Selection of 100 Paraphrase Candidates for detailed Annotation.} The sample was selected based on assigned categories during paraphrase candidate annotation. Categories within Paraphrase and Non-Paraphrase can overlap. We display  ``accuracy'' w.r.t. first author annotations.}
        \label{tab:100PC-Selection}
    \end{table}

        \begin{figure}[t]
            \centering
            \includegraphics[width=\columnwidth]{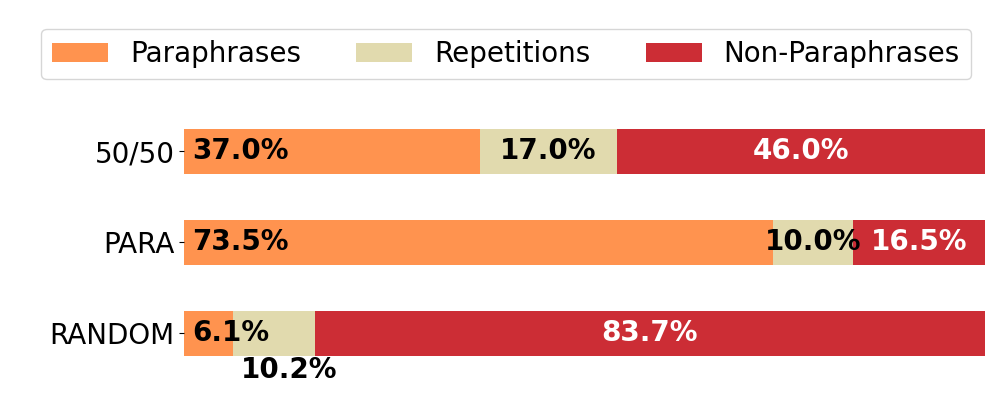}
            \caption{\textbf{Distribution of Labels by Lead Author.} We display the estimated number of (non-)paraphrases from the lead author annotations for the random subsample (RANDOM), the BALANCED sample and the wider paraphrase variety sample (PARA). Note, RANDOM consists of 100 elements, however only 98 are included in this statistic here (leading to numbers like 6.1). 2 pairs were not classified by the lead author because they were too ambiguous or were missing context information to reach a decision. We exclude such pairs in all other samples.}
            \label{app:label-dist}
        \end{figure}    

    \subsection{Paraphrase Candidate Selection} \label{app:selection}

    Based on the lead author classifications into paraphrase, non-paraphrase and repetition, we build three datasets for annotation (main paper \S\ref{sec:candidate-selection}). We display the first author classification distribution for the three datasets in Figure \ref{app:label-dist}.

    \textbf{BALANCED.} The BALANCED set is a sample of 100 (guest, host)-pairs that were randomly sampled based on the first batch of lead author annotations (\S \ref{app:paraphrase-candidates}).  We had additional lead author labels available for this set, see Table \ref{tab:100PC-Selection} for the distribution of these on the BALANCED set. Constraints were 50 paraphrases and 50 non-paraphrases. In order to include more complex cases, we sampled more difficult than unrelated non paraphrase pairs and we limited the number of repetition paraphrases (51\% of paraphrases are repetitions in the full batch, but only 33\% of paraphrases in BALANCED are repetitions). Due to a sampling error, we ended up with a 46/56 split. Later, we calculate the majority vote of the 20--21 annotations per (guest, host)-pair on this set, and then evaluate it by comparing it against the lead author classification, see  ``acc.'' column.

    \textbf{RANDOM.} The random set is a sample of 100 (guest, host)-pairs that was uniformly sampled from the second batch of lead author annotations (\S \ref{app:paraphrase-candidates}). 

    \textbf{PARA.} After selecting the RANDOM set, the PARA set of 400 (guest, host)-pairs was sampled to reach a specified total 350 paraphrases and 150 non-paraphrases together with the RANDOM set.\footnote{
        RANDOM and PARA were undergoing annotation together in a second annotation round, after BALANCED had already been annotated. The aim was to reach a higher distribution of paraphrases in our released dataset. The 350/150 split was somewhat arbitrary. It could have easily been 400/100 or 300/200 as well. 
    } The PARA set was selected to make the total number of non-repetition paraphrases together with RANDOM reach 300, while limiting the amount of repetition paraphrases to 50. Conversely, non-paraphrases were sampled to add up to 150. This led to 66 non-paraphrases and 334 paraphrases being sampled for the PARA set.

    \newpage

    \begin{table*}[t]
            \centering \footnotesize
             \begin{tabular}{%
                p{9mm} | p{115mm} | p{23mm}} 
             \toprule
              \textbf{Who?} &  \textbf{Example} & \textbf{see Instructions} \\ 
            \midrule
            \vspace{-2\baselineskip}\textbf{Self-Dis\-a\-gree\-ment} &
            \parbox{115mm}{
                    Guest: [..] So \uline{there was a consensus organization last year that people from genetics and ethics law got together and said, in theory, it should be acceptable to try this in human beings. The question will be, how much safety and evidence do we have to have from animal models before we say it's acceptable.}
                    \\ 
                    Host: When it comes to this issue, let's face it, \uline{while there are the concerns here in the United States}, it's happening in other countries.} \vspace{-\baselineskip}
                    & \vspace{-3\baselineskip}
                        \textbf{(C)} distinguish paraphrases from inferences, conclusions or ``just'' highly related utterances \\
                \midrule
                \multirow{14}{9mm}{\textbf{Lab Members}}  
                    & \parbox{115mm}{
                        Guest:  Hey, it's going to be a long and a long week, and we're \uline{going to} use every single minute of it to make sure that Americans know that Al Gore and Joe Lieberman are fighting for working families, right here in Los Angeles and across America. \\
                        Host:  And are you guys ready \uline{to go}? }
                        & \multirow{3}{23mm}{
                            \textbf{(P)} short subselections of tokens might be ``paraphrases'' that do not adequately represent the content of the guest's utterance}
                            \\
                    \cmidrule{2-2}
                    & \parbox{115mm}{
                        Guest: [...] There are militant groups out there firing against \uline{the military}. And we just - we really don't know who is whom. \\
                        Host: Why did \uline{the army} decide today to move in and clear out the camp?} \\ 
                    \cmidrule{2-2} 
                    & \parbox{115mm}{
                        Guest: Police have indicated that they have been getting cooperation from the people involved, of course, they are looking at all of \uline{her personal relationships} to see if there were any problems there. [...] \\
                        Host:  Well what have \uline{family members} told you? I know you've talked to various \uline{members of her family}. I understand she never missed her shifts at the restaurant where she worked. [...]}
                        \\
                    \cmidrule{2-3}
                    & \parbox{115mm}{
                        Guest: \uline{Yes, it is, all \$640,000.} \\
                        Host: \uline{That's a lot of dough.}
                    } & \vspace{-\baselineskip}
                        \textbf{(CD)} emphasize situational aspect to annotators, \textbf{(H)} ask for token-level accuracy of highlights \\
                                \midrule
                \multirow{3}{9mm}{\textbf{Prolific Annotators}} &
                \parbox{115mm}{ %
                        Guest: [...]  He was an employee that worked downtown Cleveland and saw it fall out of the armored car carrier, \uline{and pick it up, and took it, and placed it in his car}. \\
						Host: And \uline{he's been holding it ever since?}} %
                        & %
                        similar to \textbf{(C)} %
                        \\
                    \cmidrule{2-3}
                    & \parbox{115mm}{
                        Guest: [...]  \uline{Would I ever thought that this would be happening, no, it is, it's crazy? Just enjoy the moment.} \\
                        Host: [...] , Magic Johnson was saying that when he first started taking meetings with investors or with business people, \uline{they didn't take him seriously, but he thought maybe they just wanted his autograph.} [...]} 
                        & %
                         \vspace*{-2.5\baselineskip}\textbf{(AT)} use annotator screening to throw out annotators more likely to produce non-sensical pairs \\
                    \cmidrule{2-3}
                    & \parbox{115mm}{    
                        Guest: [...] they say, you, you must sue ``Fortnite'', \uline{and I'm like, ``Fortnite'', what is that? I don't even know what it is} -- \\
                        Host: \uline{So you weren't even familiar?}} 
                        & \vspace{-1.6\baselineskip}
                            \textbf{(AT)} throw out annotators that do not select obvious pairs \\
             \bottomrule
            \end{tabular}%
            \caption{\textbf{Examples of Disagreements in Paraphrase Annotation Pilots.
            }\label{ex:disagreements} All of the presented examples were \uline{highlighted} %
            by at least one annotator and selected as not showing any paraphrases at all by at least one other annotator. We show examples from three different conditions: Self-disagreement for the lead author, disagreements between volunteers/lab members and disagreements between \texttt{Prolific} annotators. These disagreements informed later training instructions: For (C), see Figure \ref{fig:annotator-training-3}; 
            for (P), see Figure \ref{fig:annotator-training-6}; for (CD), see Figure \ref{fig:annotator-training-7}; for (H), see Figure \ref{fig:annotator-training-5}; for (AT), we chose the separate training setup with attention and comprehension checks, see Figures \ref{fig:annotator-training-2}, \ref{fig:annotator-training-8} and \ref{fig:annotator-training-10}. Early on, we chose to include repetitions in our paraphrase definition since it turned out to be conceptually difficult to separate the two -- especially in a context-dependent setting (e.g., is ``You don't know.'' a repetition of ``I do not know it.'' or not?), see Figure \ref{fig:annotator-training-1}.}
        \end{table*}

    \section{Annotations} \label{app:paraphrase-annotations}

    \subsection{Development of Annotator Training.} \label{app:annotator-training-development}
    The eventual study design used in this work (see \S \ref{sec:annotation}) is the product of iterative improvement with lab members, other volunteers and \texttt{Prolific} annotators. They iterative steps can roughly be separated into: 
    
    (1) The lead author repeatedly annotated the same set of (guest, host)-pairs with a time difference of one week. See an example of early self-disagreement in Table \ref{ex:disagreements}.
    
    (2) With insights from (1) and our definition of context-dependent paraphrases, we created annotator instructions. We iteratively improved instructions while testing them with volunteers, lab members and \texttt{Prolific} crowd-workers. See examples of disagreements that led to changes in Table \ref{ex:disagreements}. 
    
    (3) Based on insights from (2), 
    we introduced an intermediate annotator training that explains paraphrase annotation in a ``hands-on'' way: Annotators have to correctly annotate a teaching example to get to the next page instead of just reading an instruction. As soon as the correct selection is made, an explanation is show (e.g., Figures \ref{fig:annotator-training-3} and \ref{fig:annotator-training-7}). After some testing rounds, we also require annotators to pass 2 attention (see Figure \ref{fig:annotator-training-10}) as well as 2 comprehension checks (see Figures \ref{fig:annotator-training-2} and \ref{fig:annotator-training-8}). 
    
    (4) We test the developed training on a selection of 20 (guest, host)-pairs out of which 10 were classified as clearly containing a paraphrase, and 10 as containing no paraphrase by the lead author, half of all examples we considered to be more difficult to classify (e.g., paraphrase with a low lexical overlap, non-paraphrase with a high lexical overlap).
    Two lab members reached pairwise Cohen of 0.51 after receiving training. Two newly recruited \texttt{Prolific} annotators reached average pairwise Cohen of 0.42 after going through training. Due to the inherent difficulty of the task and the good annotation quality when manually inspecting the 20 examples for each annotator, we carry on with this training setup.

    \subsection{Annotator Training.} \label{app:training}
\begin{figure}
        \centering
        \includegraphics[width=.47\textwidth]{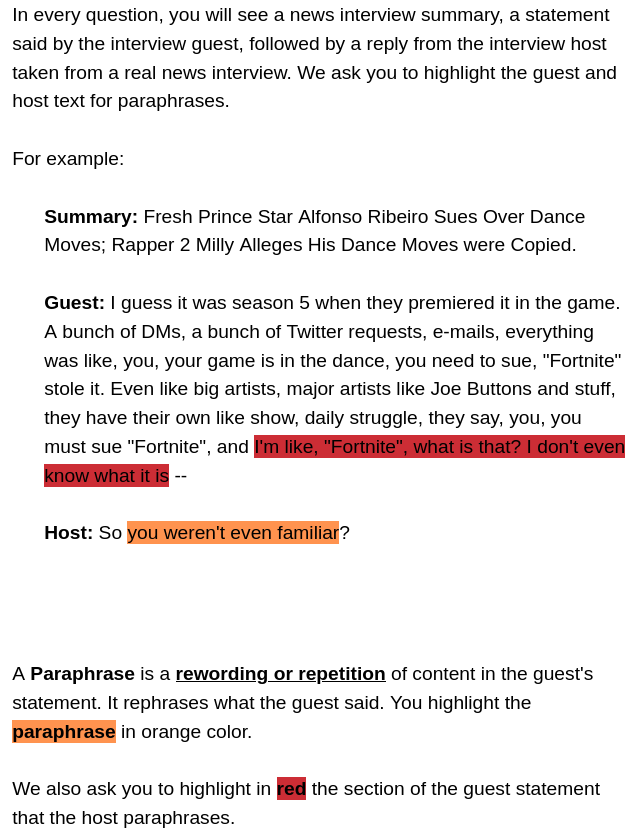}
        \caption{\textbf{Annotator Training (1).} Definition Paraphrase}
        \label{fig:annotator-training-1}
    \end{figure}

\begin{figure}
        \centering
        \includegraphics[width=.47\textwidth]{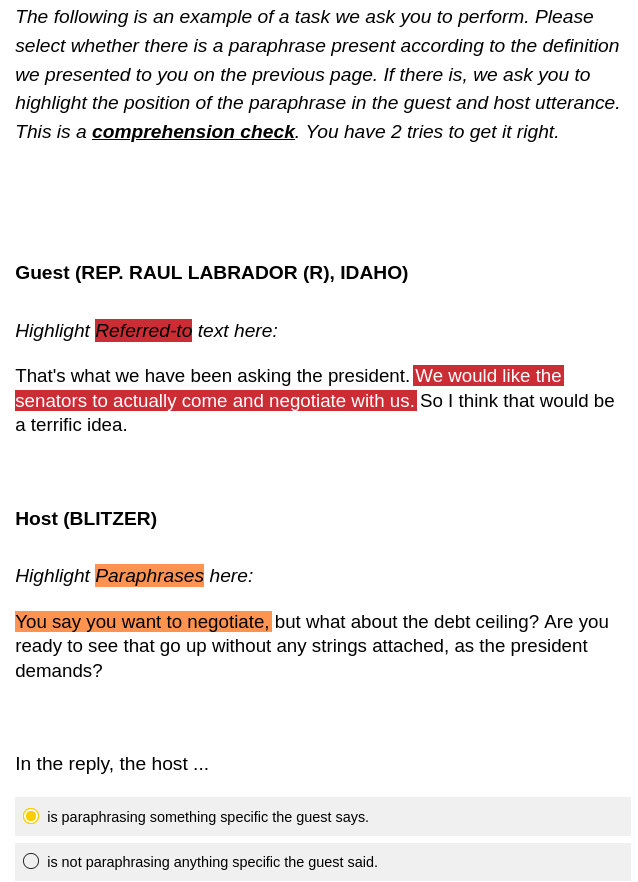}
        \caption{\textbf{Annotator Training (2).} Comprehension Check Paraphrase. Variations of the the shown highlighting are accepted.}
        \label{fig:annotator-training-2}
    \end{figure}

\begin{figure}
        \centering
        \includegraphics[width=.47\textwidth]{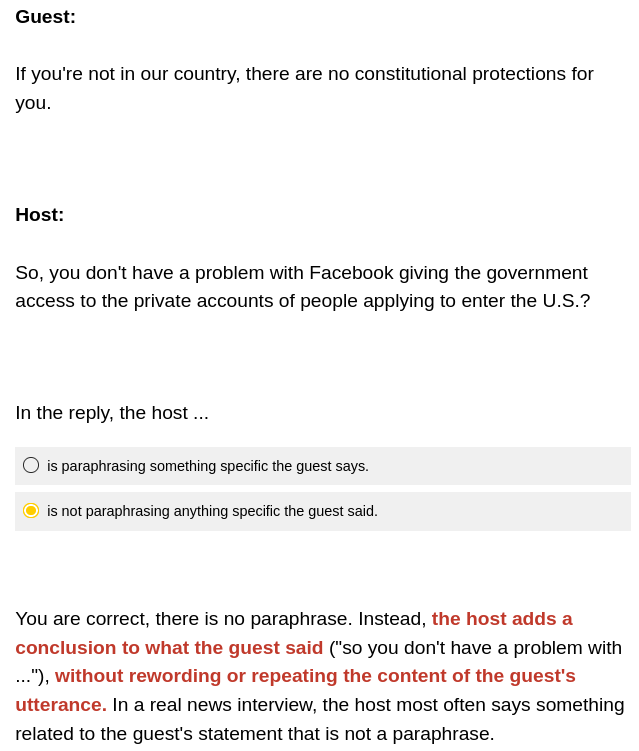}
        \caption{\textbf{Annotator Training (3).} Related but not a Paraphrase}
        \label{fig:annotator-training-3}
    \end{figure}

\begin{figure}
        \centering
        \includegraphics[width=.47\textwidth]{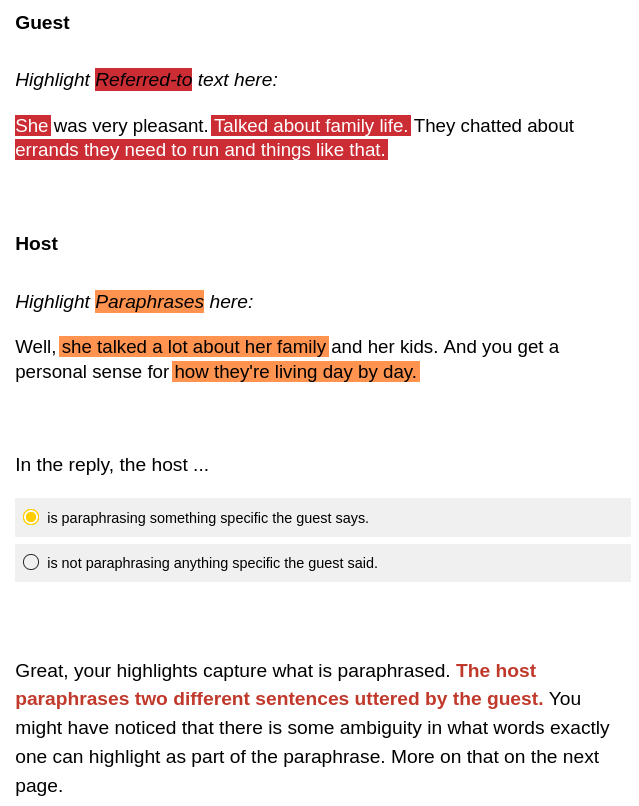}
        \caption{\textbf{Annotator Training (4).} Multiple Sentences.}
        \label{fig:annotator-training-4}
    \end{figure}

\begin{figure}
        \centering
        \includegraphics[width=.47\textwidth]{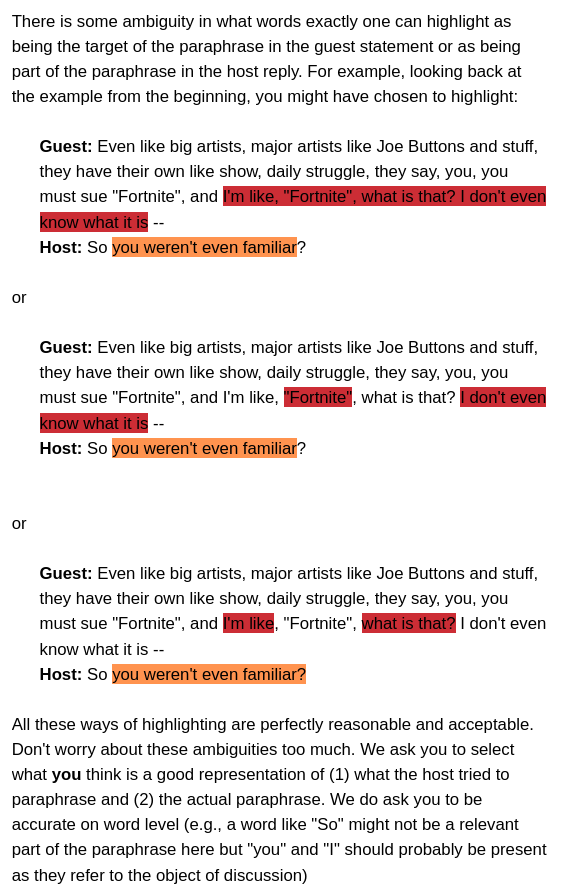}
        \caption{\textbf{Annotator Training (5).} Highlighting}
        \label{fig:annotator-training-5}
    \end{figure}

\begin{figure}
        \centering
        \includegraphics[width=.47\textwidth]{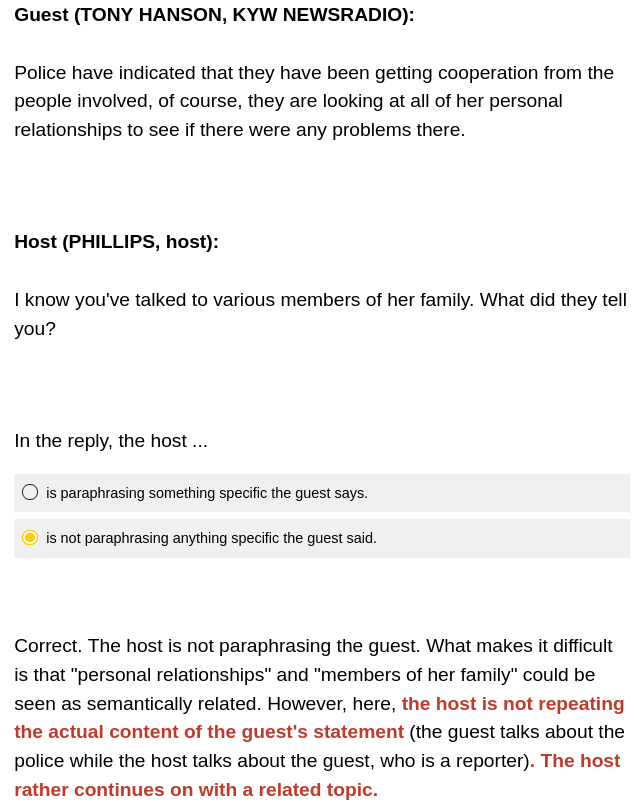}
        \caption{\textbf{Annotator Training (6).} Partial vs actual paraphrase}
        \label{fig:annotator-training-6}
    \end{figure}

\begin{figure}
        \centering
        \includegraphics[width=.47\textwidth]{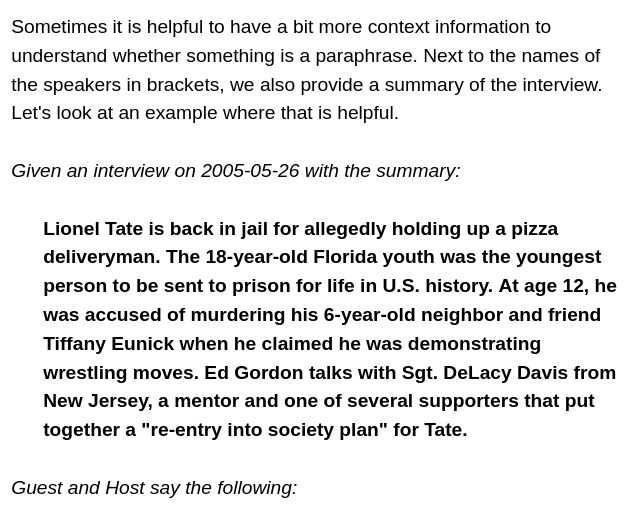}
        \includegraphics[width=.47\textwidth]{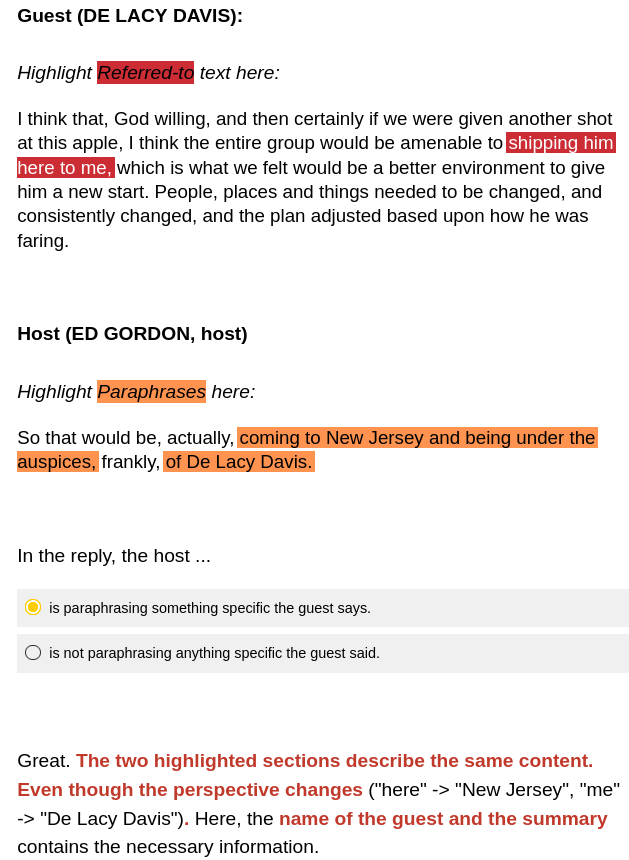}
        \caption{\textbf{Annotator Training (7).} Using context information}
        \label{fig:annotator-training-7}
    \end{figure}

\begin{figure}
        \centering
        \includegraphics[width=.47\textwidth]{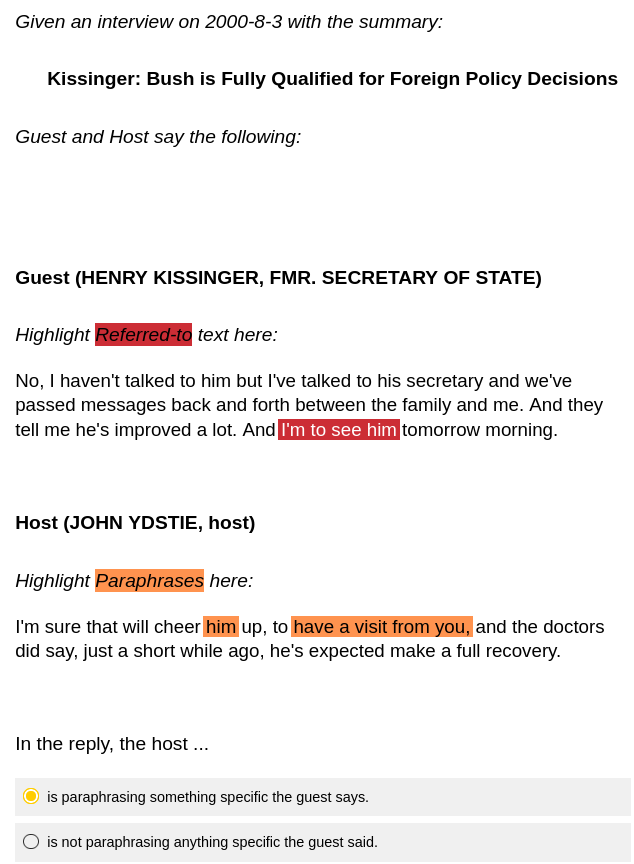}
        \caption{\textbf{Annotator Training (8).} Example of an accepted answer for the comprehension check at the end. Only annotators who highlighted similar spans are admitted to annotate unseen instances. Some of the admitted annotators additionally selected the pair ``he's improved a lot'' and ``he's expected to make a full recovery''.}
        \label{fig:annotator-training-8}
    \end{figure}

\begin{figure}
        \centering
        \includegraphics[width=.47\textwidth]{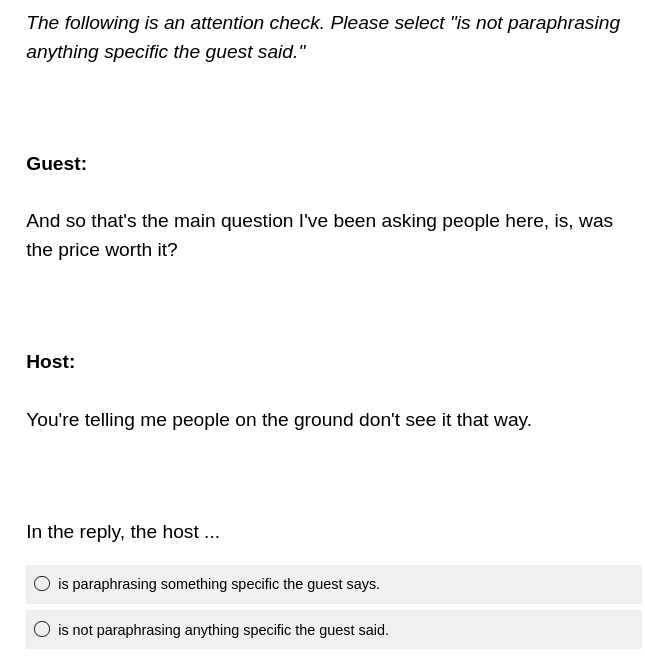}
        \includegraphics[width=.47\textwidth]{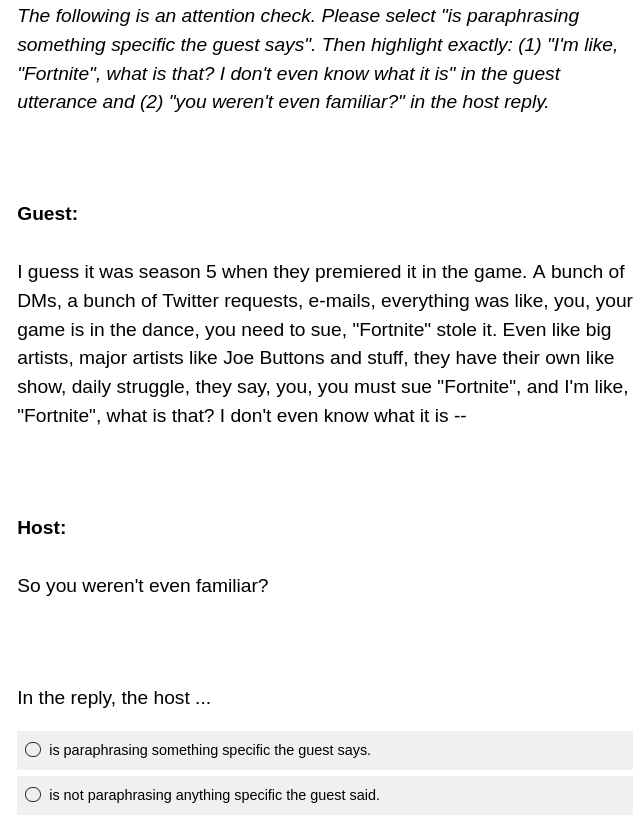}
        \caption{\textbf{Annotator Training (10).} Two attention checks shown at different times during training.}
        \label{fig:annotator-training-10}
    \end{figure}

\begin{figure*}
        \centering
        \includegraphics[width=.8\textwidth]{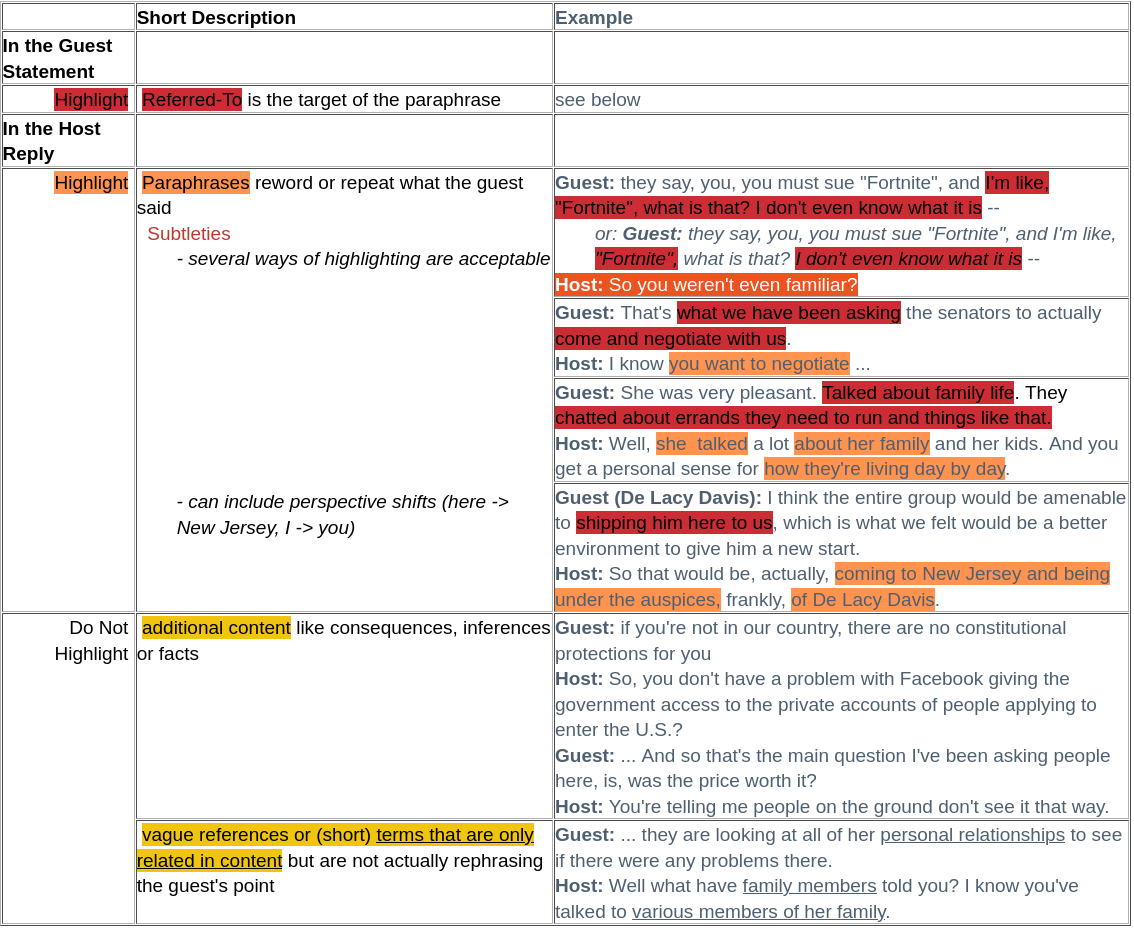}
        \caption{\textbf{Annotator Training (9).} Overview Table shown to annotators}
        \label{fig:annotator-training-9}
    \end{figure*}

    We train participants to recognize paraphrases (see Figure \ref{fig:annotator-training-1}--\ref{fig:annotator-training-9} for the instructions they received). We presented (guest, host)-pairs with their MediaSum summaries, the date of the interview and the interviewer names for context.\footnote{
        The additional information of summary, date and speaker names increased reported understanding of context and eased difficulty of the task in pilot studies among lab members.
    }  %
    Participants were only admitted to the paraphrase annotation if they passed two attention checks (see Figure \ref{fig:annotator-training-10}) and two comprehension checks (see Figure \ref{fig:annotator-training-2} and \ref{fig:annotator-training-8}). %
    
    \textbf{Comprehension Checks.} Similar to examples in Table \ref{tab:CP-examples}, they are presented with a clear paraphrase pair (App. Figure \ref{fig:annotator-training-2}) and a less obvious context-dependent paraphrase pair (App. Figure \ref{fig:annotator-training-8}) that they have to classify as a paraphrase. Additionally, they are only allowed to highlight the text spans that are a part of the paraphrase. 
    
    \textbf{Training Stats.} Of the initial 347 \texttt{Prolific} annotators who started the training, 
            95 aborted the study without giving a reason\footnote{Usually quickly, we assume that they did not want to take part in a multi-part study or did not like the task itself.} and 
            126 were excluded from further studies because they failed at least one comprehension (29\%) or attention check (24\%) during training. 
        Since annotators can perform annotations after training over a span of several days, we further exclude single annotation sessions, where the annotator fails any of two attention checks. %

        \clearpage
        \pagebreak

    
    \clearpage
    \newpage
    \pagebreak

    \subsection{Annotation After Training.} 
\begin{figure}[t]
        \centering
        \includegraphics[width=.47\textwidth]{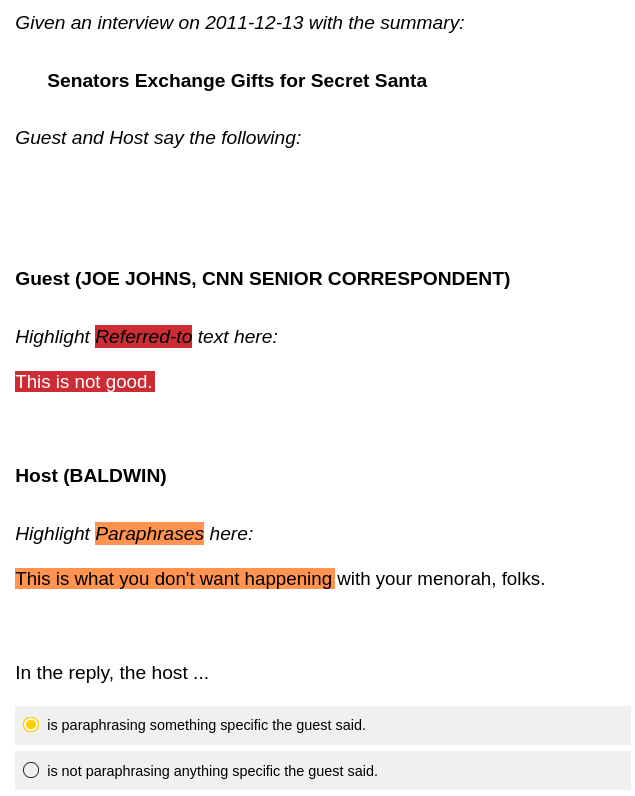}
        \caption{\textbf{Interface for highlighting categories.} Annotators are asked to highlight the categories on word level.}
        \label{fig:interface-highlighting}
    \end{figure}

    Next, the trained annotators were asked to highlight paraphrases. See Figure \ref{fig:interface-highlighting} for an example of the annotation interface. Annotators had access to a summary of their training at all times, see Figure \ref{fig:annotator-training-9}. We again included two attention checks. Answers failing either attention check are removed from the dataset.

            \begin{figure*}[hbt!] 
            \centering
            \begin{subfigure}{0.48\textwidth}
                 \centering
                 \includegraphics[width=1\textwidth]{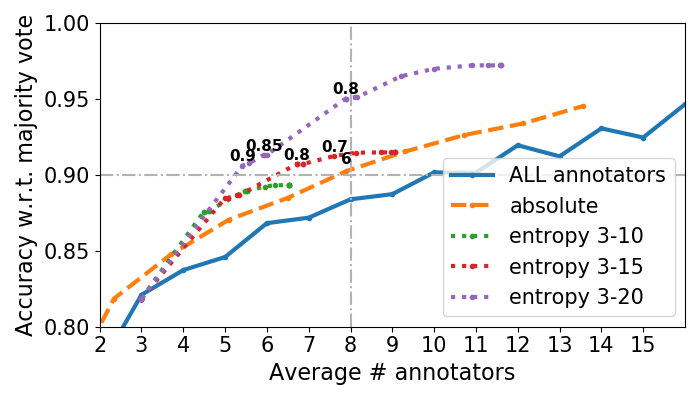}
                 \caption{Accuracy w.r.t. 20 annotators}
                 \label{subfig:accuracy}
             \end{subfigure}
             \begin{subfigure}{0.48\textwidth}
                 \centering
                 \includegraphics[width=1\textwidth]{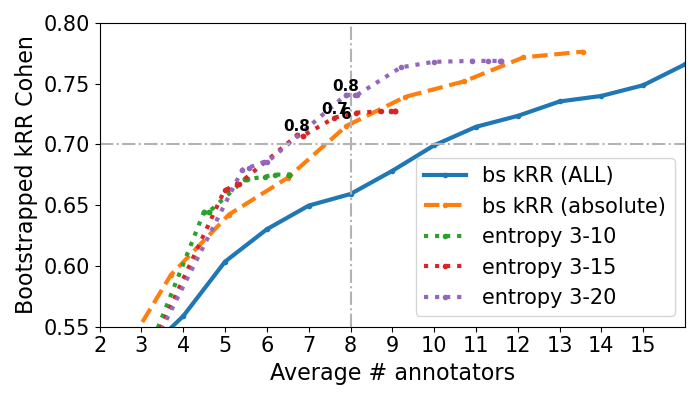}
                 \caption{kRR}
                 \label{subfig:kRR}
             \end{subfigure}
             \caption{
                \textbf{Annotator Recruitment Strategies.} 
                To decide the number of annotators for a specific item, we test three different strategies: (1) using a fixed number of annotators across all items (ALL), (2) increasing the number of annotators until at least $n$ annotators agree for each item (absolute) and (3) increasing the number of annotators from 3 until the entropy is smaller than a given threshold (entropy) or a maximum of 10, 15 or 20 annotators is reached.
                We display the accuracy of the methods compared to using all 20 annotations in (\ref{subfig:accuracy}) and the reliability measure kRR depending on the average number of annotators used \citep{wong-paritosh-2022-k} in (\ref{subfig:kRR}). 
                We set a maximum average cost of 8 annotators per item and require a minimum accuracy of 90\% as well as a minimum kRR of 0.70. 
                When a strategy fulfills these requirements (i.e., falls in the upper left quadrants for (a) and (b)), we display the entropy thresholds for (3) and absolute number of annotators for (2). 
             }
             \label{fig:nbr-annotators}
        \end{figure*}

    \subsection{Annotator Allocation Strategy} \label{app:annotator-allocation}

        To the best of our knowledge, what constitutes a ``good'' number of annotators per item has not been investigated for paraphrase classification. 

        \textbf{Summary.} Based on the 20--21 annotations per item for the BALANCED set, we simulate fixed and dynamic strategies to recruit up to 20 annotations per item. We evaluate the different strategies w.r.t. closeness to the annotations of all 20--21 annotators. 
        When considering resource cost and performance trade-offs, dynamic recruitment strategies performed better than allocating a fixed number of annotators for each item.

       \textbf{Details.} We consider three different strategies for allocating annotators to an item: (1) using a fixed number for all items, (2) for each item, dynamically allocate annotators until $n$ of them agree and (3) similar to \citet{engelson-dagan-1996-minimizing}, for each item, dynamically allocate annotators until the entropy is below a given threshold $t$ or a maximum number of annotators has been allocated. 
        We simulate each of these strategies using the annotations on BALANCED. 
        We evaluate the strategies on (a) cost, i.e., the average number of annotators per item and (b) performance via (i) the overlap between the full 20 annotator majority vote (i.e., we assume this is the best possible result) and the predicted majority vote for the considered strategy and (ii)  k-rater-reliability \citep{wong-paritosh-2022-k} --- a measure to compare the agreement between aggregated votes. Note, for the dynamic setup we change the original calculation of kRR \citep{wong-paritosh-2022-k} by dynamically recruiting more or less annotators per item and thus aggregating the votes of a varying instead of a fixed number of annotators. 

        \textbf{Results.} See Figure \ref{fig:nbr-annotators} for the results. We selected a practical resource limit of an average 8 annotators per items and the requirement of at least 90\% accuracy with the majority vote and 0.7 kRR (dotted lines). We decide on strategy (3)  dynamically recruiting annotators (minimally 3, maximally 15) until entropy is below 0.8. Also with other min/max parameters this was a good trade-off between accuracy, kRR and average \# of annotators.
        The average number of annotators needed per item is then about 6.8. In this way, most items receive annotations from 3 annotators, while difficult ones receive up to 15.

    \subsection{Annotator Payment.} Via Prolific's internal screening system, we recruited native speakers located in the US. Payment for a survey was only withheld if annotators failed two attention checks within the same survey or when a comprehension check at the very beginning of the study was failed\footnote{Technically, in line with Prolific guidelines, we do not withhold payment but ask annotators to ``return'' their study in this case. Practically this is the same, as all annotators did return such a study when asked.} in line with Prolific guidelines.\footnote{\href{https://researcher-help.prolific.co/hc/en-gb/articles/360009223553-Prolific-s-Attention-and-Comprehension-Check-Policy}{Prolific Attention and Comprehension Check Policy}}
    Across all \texttt{Prolific} studies performed for this work (including pilots), we payed participants a median of $8.98\pounds/h\approx11.41\$/h$\footnote{on March 20th 2024}
    which is above federal minimum wage in the US.\footnote{Federal minimum wage in the US is $\$7.25/h\approx5.71\pounds/h$ according to \url{https://www.dol.gov/agencies/whd/minimum-wage} on March 20th 2024} %
    
    \begin{figure*}[!htb] 
            \begin{subfigure}{0.4\textwidth}
                 \centering
                 \includegraphics[width=1\textwidth]{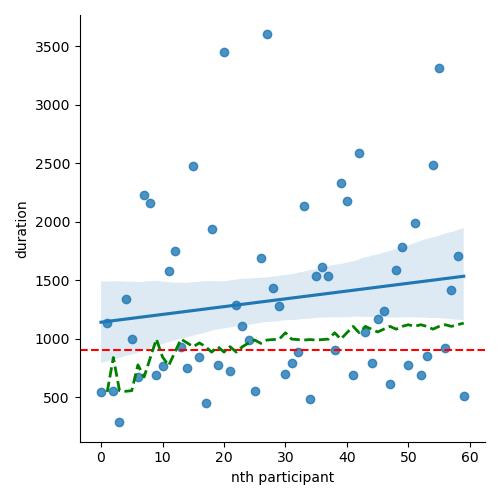}
                 \caption{Duration}
                 \label{subfig:duration-training}
             \end{subfigure}
             \begin{subfigure}      {0.5\textwidth}
                 \centering
                 \includegraphics[width=1\textwidth]{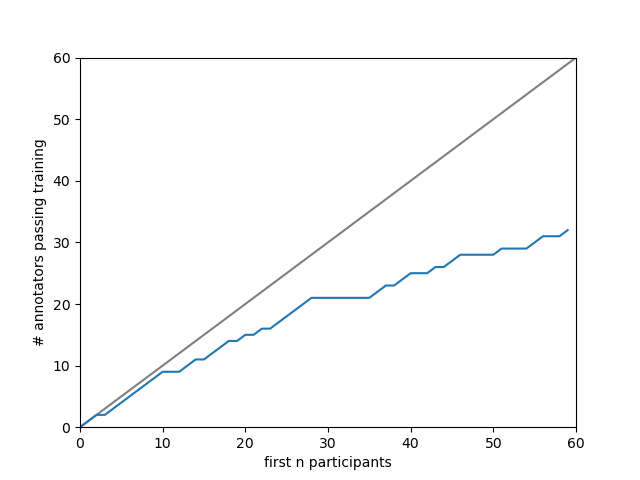}
                 \caption{Quality Checks Passed}
                 \label{subfig:quality-checks-training}
             \end{subfigure}
             \caption{\textbf{On BALANCED, later training sessions take longer and pass fewer quality checks.} \label{fig:later-participants-worse}
                In \ref{subfig:duration-training}, we display the seconds the nth annotator needs to go through the training session. The annotators are ordered according to the dates they completed training. Annotations were distributed across 6 different days in June 2023. The green line represents the median duration time of the first n participants. The red line displays the initially estimated completion time of 900 seconds according to pilot studies. The blue line is a linear regression estimate of the duration and it's 95\% confidence interval. On average, participants participating on a later date need more time to finish. In \ref{subfig:quality-checks-training}, we display the summed number of the first n participants that passed the quality checks during training. The grey line represents the angle bisector, i.e., if every participant would pass all quality checks. Later participants are less likely to pass the quality checks. %
             }
        \end{figure*}

     \subsection{Varying Annotator Behavior over Time.}
    For the BALANCED set, we performed separate training and annotation rounds. See Figure \ref{fig:later-participants-worse} for the completion times and share of passed quality checks of \texttt{Prolific} annotators in the training session. Participants that were recruited later performed worse: they pass less quality checks and need more time. 
    This effect was noticeable but it is not quite clear to us why this happens. We recruit all participants at once for later studies and not iteratively as for the BALANCED set, to avoid effects that have to do with study age. The effect on the quality of the released annotations should be minimal as we discard annotators that do not pass our quality checks. It does have an effect on the pay per hour for our participants, which we had initially estimated to be much higher. 

    \subsection{Intra-Annotator Annotations Quality} \label{app:intra-annotator-annotations}
    
    We manually randomly sample ten annotators (with anonymized PROLIFIC ids 60, 6, 86, 84, 47, 31, 68, 88, 41, 92) and analyze  42 of their annotatations. Nine annotators consistently provide plausible annotations, while the other annotator chooses “not a paraphrase” a few times too often. We also noticed some other annotator-specific tendencies, for example, one annotator might tend to highlight fewer words, more words or prefer exact lexical matches.

    \subsection{Anonymization} We replace all \texttt{Prolific} annotator IDs with non-identifiable IDs. We only make the non-identifiable IDs public.

    \clearpage
    \pagebreak
    \newpage

    \clearpage
    \newpage
    \pagebreak
    
    \section{Modeling} 

    \subsection{In-Context Learning} 
    
    \label{app:icl-models}
    
     \paragraph{Models.} We provide the Huggingface URLs to our used models. Vicuna 7B: \url{https://huggingface.co/lmsys/vicuna-7b-v1.5}, Mistral 7B Instruct v0.2: \url{https://huggingface.co/mistralai/Mistral-7B-Instruct-v0.2}, Openchat: \url{https://huggingface.co/openchat/openchat-3.5-0106}, Gemma 7B: \url{https://huggingface.co/google/gemma-7b-it}, Mixtral 8x7B Instruct v0.1: \url{https://huggingface.co/mistralai/Mixtral-8x7B-Instruct-v0.1}, Llama 7B: \url{https://huggingface.co/meta-llama/Llama-2-7b-hf} and Llama 70B: \url{https://huggingface.co/meta-llama/Llama-2-70b-hf}.
    
    \label{app:prompting}

    \paragraph{Prompt.} We use a few-shot prompt that is close to the original annotator training and instructions, see Figure~\ref{fig:prompt-instruction}. We use chain-of-thought like explanations, i.e., always starting with ``Let's think step by step.'' and ending with ``Therefore, the answer is'', \citep{kojima2022large} and a few-shot setup showing all 8 examples showed to annotators during training (Figures \ref{fig:annotator-training-1}--\ref{fig:annotator-training-10}). 
    For \texttt{GPT-4}, we use a temperature of 1, self-consistency through prompting the model 3 times \citep{wang2022self} and the default top\_p nucleus sampling value of 1, a maximum of new tokens to 512.
    For all the huggingface models, we use a temperature of 1, self-consistency through prompting the model 10 times (only 3 times for \texttt{Lllama 70B} due to resource limits) and a top\_k sampling of the top 10 tokens, a maximum of new tokens of 400 for all other models.
    Note, there are many more prompts and choices we could have tried that are out-of-the scope of this work. Further steps could have included separating the classification and highlighting task, experimenting with further phrasings and so on. We leave this to future work.

    \begin{figure*}[h]
    \centering
    \tiny
    \begin{mdframed}
    \begin{lstlisting}[breaklines]
A Paraphrase is a rewording or repetition of content in the guest's statement. It rephrases what the guest said.
Given an interview on - with the summary: Fresh Prince Star Alfonso Ribeiro Sues Over Dance Moves; Rapper 2 Milly Alleges His Dance Moves were Copied.
Guest and Host say the following:
Guest (TERRENCE FERGUSON, RAPPER): I guess it was season 5 when they premiered it in the game. A bunch of DMs, a bunch of Twitter requests, e-mails, everything was like, you, your game is in the dance, you need to sue, "Fortnite" stole it. Even like big artists, major artists like Joe Buttons and stuff, they have their own like show, daily struggle, they say, you, you must sue "Fortnite", and I'm like, "Fortnite", what is that? I don't even know what it is --
Host (QUEST): So you weren't even familiar?
In the reply, does the host paraphrase something specific the guest says?

Explanation: Let's think step by step.
Terrence Ferguson says at the end of his turn that he didn't know Fortnite.
Quest, the host of the interview, repeats that the guest doesn't know Fortnite.
So they both say that the guest didn't know Fortnite. Therefore, the answer is yes, the host is paraphrasing the guest.
Verbatim Quote Guest: "I'm like, "Fortnite", what is that?  I don't even know what it is"
Verbatim Quote Host: "you weren't even familiar?"
Classification: Yes.


Given an interview on 2013-10-1 with the summary: ...
Guest and Host say the following:
Guest (REP. RAUL LABRADOR (R), IDAHO): ...
Host (BLITZER): ...
In the reply, does the host paraphrase something specific the guest says?

Explanation: Let's think step by step. EXPLANATION Therefore, the answer is yes, host is paraphrasing the guest.
Verbatim Quote Guest: "We would like the senators to actually come and negotiate with us."
Verbatim Quote Host: "you want to negotiate"
Classification: Yes.


ITEM

Explanation: ...
Verbatim Quote Guest: None.
Verbatim Quote Host: None.
Classification: No.

ITEM

Explanation: ...
Verbatim Quote Guest: "She" "Talked about family life." "errands they need to run and things like that."
Verbatim Quote Host: "she talked" "about her family and her kids." "how they're living day by day."
Classification: Yes.

ITEM

Explanation: ...
Verbatim Quote Guest: None.
Verbatim Quote Host: None.
Classification: No.


ITEM

Explanation: ...
Verbatim Quote Guest: None.
Verbatim Quote Host: None.
Classification: No.


ITEM

Explanation: ...
Verbatim Quote Guest: "shipping him here to me"
Verbatim Quote Host: "coming to New Jersey and being under the auspices" "of De Lacy Davis."
Classification: Yes.


ITEM

Explanation: ...
Verbatim Quote Guest: "I'm to see him."
Verbatim Quote Host: "him" "have a visit from you"
Classification: Yes.


Given an interview on DATE with the summary: SUMMARY
Guest and Host say the following:
Guest (NAME): UTTERANCE
Host (NAME): UTTERANCE
Explanation: Let's think step by step.

    \end{lstlisting}

    \end{mdframed}
    \caption{\textbf{Prompt Template close to Annotator Instructions} The used prompt template is based closely on our annotator training and instructions. Phrasings were adapted to match the prompt-setting but kept the same where possible. See the full prompt in our Github Repository. \label{fig:prompt-instruction}}
    \end{figure*}

        \begin{table}[t]
        \centering \small
         \begin{tabular}{l r | c} 
         \toprule
         
          Learning Rate & Epoch & F1 \\
        \midrule
            1e-3 & 8 & 0.61 $\pm$ 0.04 \\
            \uline{3e-3} & 8 & {0.64} $\pm$ 0.06 \\
            5e-3 & 8 & 0.52 $\pm$ 0.15 \\
                
        \midrule
            3e-3 & 4 & {0.65} $\pm$ 0.07 \\
            3e-3 & \uline{12} & \textbf{0.65} $\pm$ 0.00\\
            3e-3 & 16 & 0.60 $\pm$ 0.10\\
         \bottomrule
        \end{tabular}%
        \caption{\textbf{Hyperparameter tuning on the DEV set.} \label{fig:token-hyperparameter} 
            We train a token classifier for learning rates 1e-3, 3e-3, 5e-3 and epochs 4, 8, 12 and 16 for 3 seeds. We keep learning rate fixed at 3e-3 when varying the number of epochs and epoch fixed at 8 when varyig the learning rates. Best options of learning rate and epoch are \uline{underlined}. Best F1 score is \textbf{boldfaced}.
        }
        \end{table}    

    \subsection{Token Classification} \label{app:token-classification}

    We use settings very close to \citet{wang-etal-2022-paratag} and test different learning rates and number of epochs with 3 different seeds each. We use the "save best model" option to save the model after the epoch which yielded the best result on the dev set. For the results, see Figure \ref{fig:token-hyperparameter}. We use a learning rate of 3e-3 and 12 epochs for further modeling.

    \clearpage
    \newpage
    \subsection{Highlighting Analysis} \label{app:error-analysis}

    We compare the highlights provided by \texttt{DeBERTa AGGREGATED}\footnote{i.e., seed $202$ with F1 score of $0.76$, precision of $0.72$ and recall of $0.84$, see \url{https://huggingface.co/AnnaWegmann/Highlight-Paraphrases-in-Dialog}} and \texttt{DeBERTa ALL}\footnote{i.e., seed $201$ with F1 score of $0.72$, precision of $0.84$ and recall of $0.63$, see \url{https://huggingface.co/AnnaWegmann/Highlight-Paraphrases-in-Dialog-ALL}}  on 10 text pairs from the test set that were classified as paraphrases by both models. We provide examples in Table~\ref{tab:deberta-highlights}. \texttt{DeBERTa ALL} highlights are shorter, often more on point and arguably more consistent than \texttt{DeBERTa AGGREGATED} highlights. We also manually analyzed 10 text pairs from the test set that GPT-4 classified as paraphrases. We provide examples of \texttt{GPT-4} highlights in Table \ref{tab:gpt4-highlights}. Generally, they seem of good quality, but have the tendency to span complete sub-sentences, even if not all is relevant.

    \textbf{Hallucinations.} One of the biggest problems for in-context learning are the extractions of the highlighting from the model responses which has errors in up to 71\% of the cases in Table \ref{tab:model-results}. Most of these errors can be split into two categories: (1) inconsistent highlighting, where the model classifies a paraphrase but does not highlight text spans in both, the guest and host utterance and (2) hallucinations, where the model highlights spans that do not exist in that form in the guest or host utterance. Hallucination is more prevalent than inconsistent highlighting for \texttt{GPT-4}, where in most cases it leaves out words (e.g., ``coming back to a normal winter'' vs. ``coming back daryn to a normal winter''), in some other cases it adds or replaces words (e.g.,``he's a counterpuncher'' vs. ``he's counterpuncher''), uses morphological variation (e.g., ``you've'' vs. ``you have'') or quotes from the wrong source (e.g., from the host when considering the guest utterance). Most of these extraction errors seem to be resolvable by humans when looking at them manually, so it might be possible to address them in future work with a more advanced matching algorithm or by querying \texttt{GPT-4} until one gets a parsable response.
    When looking at the classifications by \texttt{GPT-4} they often seem plausible, even when counted as incorrect with the F1 score.

    \subsection{Computing Infrastructure}

    The fine-tuning of 18 DeBERTa token classifier, and the inference of 7 generative models took about approximately 260 GPU hours with one A100 card with 80GB RAM on a Linux computing cluster.

    We use \texttt{scikit-learn} $1.2.2$ \citep{scikit}, %
    \texttt{statsmodels} $0.14.1$ \citep{seabold2010statsmodels}  %
    and \texttt{krippendorff} $0.6.1$ \citep{castro-2017-fast-krippendorff}
    for evaluation.

        \begin{table*}[th]
            \centering \footnotesize 
             \begin{tabular}{
               c c c p{120mm}} 
             \toprule
               \texttt{AGG} & \texttt{ALL} & C &  \textbf{Shortened  Examples} \\ 
            \midrule
                \cmark & \cmark & \xmark & \parbox{120mm}{
                    \textbf{G: }
                        There are people that are in that age range where we know they're high risk, %
                        \colorlet{orange28}{orange!28.57142857142857}\sethlcolor{orange28}\hl{why }\colorlet{orange14}{orange!14.285714285714285}\sethlcolor{orange14}\hl{are they going to the}\colorlet{orange28}{orange!28.57142857142857}\hluline{orange28}{supermarket}\sethlcolor{orange28}\hl{ to}\colorlet{orange42}{orange!42.857142857142854}\sethlcolor{orange42}\hluline{orange42}{\textbf{buy} their}\hluline{orange42}{\textbf{ own groceries?}}\colorlet{orange57}{orange!57.14285714285714}\sethlcolor{orange57}\hl{ Get the community, the neighborhood }\colorlet{orange71}{orange!71.42857142857143}\sethlcolor{orange71}\hluline{orange71}{ to}\colorlet{orange85}{orange!85.71428571428571}\hluline{orange85}{ go and help them.}
                     \\ 
                    \textbf{H: }
                        \colorlet{orange28}{orange!28.57142857142857}\sethlcolor{orange28}\hl{if }\colorlet{orange42}{orange!42.857142857142854}\hluline{orange42}{you're}\colorlet{orange57}{orange!57.14285714285714}\hluline{orange57}{going to}\colorlet{orange85}{orange!85.71428571428571}\hluline{orange85}{ help somebody}\colorlet{orange71}{orange!71.42857142857143}\hluline{orange71}{by }\colorlet{orange100}{orange!100.0}\sethlcolor{orange100}\hluline{orange100}{\textbf{helping them}}\colorlet{orange85}{orange!85.71428571428571}\sethlcolor{orange85}\hluline{orange85}{ \textbf{maybe get their groceries},}  how long does the coronavirus live on surfaces?
                } \\  \midrule
                \cmark & \cmark & \cmark & \parbox{120mm}{
                    \textbf{G: }
                    \colorlet{orange0}{orange!0}\sethlcolor{orange0}\hl{
                    And people always prefer, of course, to see the pope as the principal celebrant of the mass. So that's good. That'll be tonight. And it will be his 26th mass and it will be the 40th or, rather, the 30th time that this is offered in round the world transmission. And }\hluline{orange100}{it will be } \hluline{orange100}{\textbf{my 20th time in doing it} as}\sethlcolor{orange100}\hl{ a }\hluline{orange100}{television commentator}\colorlet{orange33}{orange!33.33333333333333}\sethlcolor{orange33}\hl{ from Rome} \colorlet{orange0}{orange!0}\sethlcolor{orange0}\hl{so. } 
                     \\ 
                    \textbf{H: }
                    \colorlet{orange0}{orange!0}\sethlcolor{orange0}\hl{Yes, } \colorlet{orange100}{orange!100.0}\sethlcolor{orange100}\hluline{orange100}{\textbf{you've been doing this for a while} now.} 
                    } \\ \midrule
                \cmark & \cmark & \cmark & \parbox{120mm}{
                    \textbf{G: }
                        \colorlet{orange0}{orange!0}\sethlcolor{orange0}\hl{Well, what happened was we finally waved down a Coast Guard helicopter. And what they were looking for were people with disabilities and medical conditions, which none of us really had. They didn't lift any of us into the helicopter or anything. What they told us was to  }\uline{basically }\colorlet{orange75}{orange!75}\hluline{orange75}{ \textbf{walk}}\colorlet{orange70}{orange!70}\sethlcolor{orange70}\hl{out}\colorlet{orange65}{orange!65}\sethlcolor{orange65}\hl{ of }\hluline{orange65}{our}\sethlcolor{orange65}\hl{ house, }\colorlet{orange55}{orange!55}\sethlcolor{orange55}\hluline{orange55}{up }\hluline{orange55}{ \textbf{the}}\hl{  \textbf{street},}\hluline{orange65}{ trying to}\colorlet{orange80}{orange!80}\hluline{orange80}{\textbf{fight against the}}\colorlet{orange85}{orange!85}\hluline{orange85}{\textbf{current} }\colorlet{orange50}{orange!50}\sethlcolor{orange50}\hluline{orange50}{\textbf{that was}}\hl{ going }\hluline{orange50}{the}\hl{\textbf{opposite} way } of where \uline{we} needed to go.
                     \\ 
                    \textbf{H: }
                        \colorlet{orange0}{orange!0}\sethlcolor{orange0}\hl{So }\hluline{orange65}{\textbf{you} }\colorlet{orange95}{orange!95}\hluline{orange95}{\textbf{walked through}}\hluline{orange100}{\textbf{ that current}} \uline{\textbf{to get to the higher ground}} or get to a drier spot? 
                    } \\ \midrule
                \cmark & \xmark & \xmark & 
                    \parbox{120mm}{
                            \textbf{G: } \colorlet{orange20}{orange!20}\sethlcolor{orange20}\hl{They've now spent \$6 million on this} \hluline{orange20}{Benghazi} \sethlcolor{orange20}\hl{investigation. They keep coming up with more and more interviews.}
                             \\ 
                            \textbf{H: } \sethlcolor{orange20}\hl{On Benghazi, Trey Gowdy now says}\hluline{orange20}{your committee has interviewed 75 witnesses.}
                    } \\ 
             \bottomrule
        \end{tabular}
        \caption{\textbf{\texttt{DeBERTa ALL} vs \texttt{DeBERTa AGGREGATED} highlights.}\label{tab:deberta-highlights} Paraphrase highlights predicted by the best \textbf{\texttt{DeBERTa ALL}} (i.e., seed 201 with F1 score of 0.72) and the best \uline{\texttt{DeBERTa AGG}} model (i.e., seed 202 F1 score of 0.76, same as in the main paper). Even though \texttt{DeBERTa AGG} gets better F1 scores on classification, the \texttt{DeBERTa ALL} highlights are arguably more on point. For comparison, we also display the human \sethlcolor{orange100}\hl{highlights} if they exist. Note, highlights can exist even if the crowd majority vote did not predict a paraphrase.
        }
        \end{table*}

        \begin{table*}[th]
            \centering \footnotesize 
             \begin{tabular}{
               c c p{130mm}} 
             \toprule
               \texttt{GPT-4} & C &  \textbf{Shortened  Examples} \\ 
            \midrule
                \cmark & \xmark & \parbox{130mm}{
                            \textbf{G: } \uline{We also want to see what connections exist between pardons and potential gifts to the Clinton Library.}
                             \\ 
                            \textbf{H: } Congressman, short of, though, having a thank-you note attached to a check that went to the \uline{Clinton Library, what is it exactly that is going to prove that there was a quid pro quo}, that these pardons were actually bought?
                    } \\ \midrule
                \cmark & \xmark & \parbox{130mm}{
                         \textbf{G: } 
                        \colorlet{orange20}{orange!20}\sethlcolor{orange20}\hl{They've now spent \$6 million on this Benghazi investigation.} \hluline{orange20}{They keep coming up with} \hluline{orange20}{more and more interviews.}
                                  \\ 
                        \textbf{H: } \sethlcolor{orange20}\hl{On Benghazi, Trey Gowdy now says} \hluline{orange20}{your committee has interviewed 75 witnesses.} 
                    } \\  \midrule
                \cmark & \cmark &
                    \parbox{130mm}{
                    \textbf{G: }
                        [Trump]\hluline{orange100}{ is appointing very young judges.} 
                     \\ 
                    \textbf{H: }
                    [...]
                    \colorlet{orange66}{orange!66.66666666666666}\sethlcolor{orange66}
                    \hluline{orange66}{if you're 50-plus, you're probably too  old for the Trump Administration to be seriously } \hluline{orange66}{considered for a district court 
                    judgeship.}
                } \\ 
             \bottomrule
        \end{tabular}
        \caption{\textbf{\texttt{GPT-4} highlights.} \label{tab:gpt4-highlights} Paraphrase highlights predicted by \uline{\texttt{GPT-4}}. For comparison, we also display the human \sethlcolor{orange100}\hl{highlights} if they exist. Note, highlights can exist even if the crowd majority vote did not predict a paraphrase.
        }
        \end{table*}

    \section{Use of AI Assistants}

    We used ChatGPT and GitHub Copilot for coding, to look up commands and sporadically to generate functions. Generated functions are marked in our code. Generated functions were tested w.r.t. expected behavior. We did not use AI assistants for writing.

\end{document}